\documentclass[10pt,twocolumn,letterpaper]{article}

\usepackage{cvpr}              

\usepackage{graphicx}
\usepackage{amsmath}
\usepackage{amssymb}
\usepackage{algorithm}
\usepackage{algorithmic}
\usepackage{multirow}
\usepackage{booktabs}
\usepackage{bbding}

\usepackage[pagebackref,breaklinks,colorlinks]{hyperref}

\usepackage[capitalize]{cleveref}
\crefname{section}{Sec.}{Secs.}
\Crefname{section}{Section}{Sections}
\Crefname{table}{Table}{Tables}
\crefname{table}{Tab.}{Tabs.}

\def\cvprPaperID{2958} 
\def\confName{CVPR}
\def\confYear{2022}

\begin{document}

\title{StrokeNet: Stroke Assisted and Hierarchical Graph Reasoning Networks}

\author{Lei Li\\
Tsinghua University\\
Haidian, Beijing, China\\
\and
Kai Fan\thanks{Corresponding author: interfk@gmail.com}\\
Alibaba Group (U.S.) Inc.\\
Sunnyvale, CA, USA\\
\and
Chun Yuan\\
Tsinghua SIGS\\
Nanshan, Shenzhen, China
}
\maketitle

\begin{abstract}
   Scene text detection is still a challenging task, as there may be extremely small or low-resolution strokes, and close or arbitrary-shaped texts. In this paper, StrokeNet is proposed to effectively detect the texts by capturing the fine-grained strokes, and infer structural relations between the hierarchical representation in the graph. Different from existing approaches that represent the text area by a series of points or rectangular boxes, we directly localize strokes of each text instance through Stroke Assisted Prediction Network (\textbf{SAPN}). Besides, Hierarchical Relation Graph Network (\textbf{HRGN}) is adopted to perform relational reasoning and predict the likelihood of linkages, effectively splitting the close text instances and grouping node classification results into arbitrary-shaped text region. We introduce a novel dataset with stroke-level annotations, namely \textbf{SynthStroke}, for offline pre-training of our model. Experiments on wide-ranging benchmarks verify the State-of-the-Art performance of our method. Our dataset and code will be available.
\end{abstract}

\section{Introduction}
Scene text detection in the wild, as a fundamental task in the computer vision field, has been widely applied in numerous applications, such as autonomous driving, document analysis and image understanding. The goal of text detection is to label each text instance with a bounding box from input images. Current leading approaches are mainly extended from the object detection or segmentation frameworks, which could be summarized into regression-based methods and segmentation-based methods, respectively. 

\begin{figure}[h]
\begin{center}
\includegraphics[width=.4\textwidth]{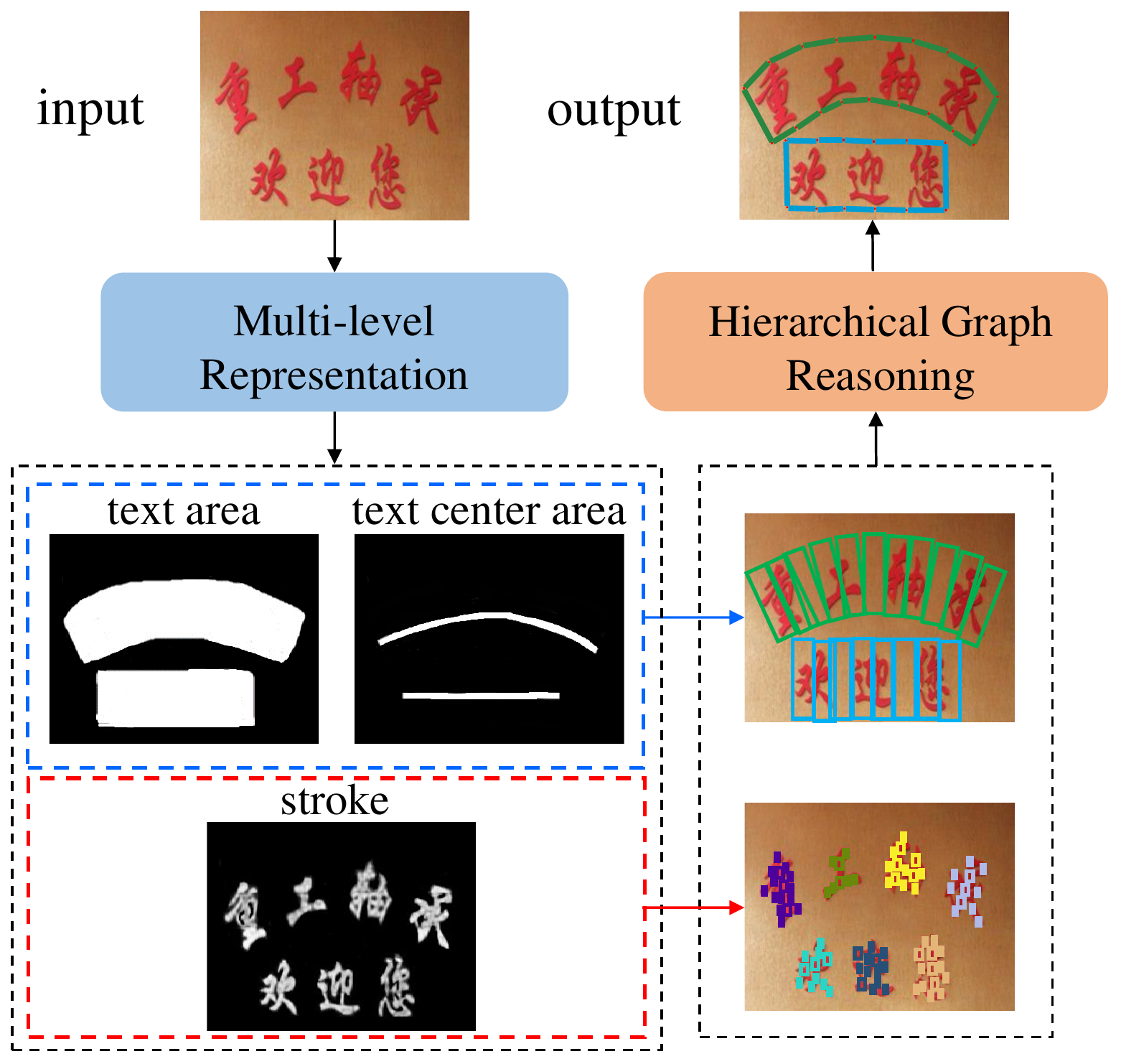}
\caption{The process of detecting texts in our StrokeNet.}
\label{new_s}
\end{center}
\end{figure}
However, they may suffer in more difficult cases. First, 
existing methods are commonly not adept in capturing the fine-grained \textbf{strokes, which are the character parts in each text-level bounding box} and play an important role in the representation of the text area. 
Second, detecting only from a text-level perspective is difficult to separate the text instances close to each other~\cite{deng2018pixellink,wang2019shape}.
Third, regression-based methods often fail to localize the text instance with arbitrary shapes~\cite{zhou2017east}, while segmentation-based methods rely on heavy post-processing~\cite{long2018textsnake} to compose the predicted regions into final text instances. 

In addition, some real-world applications, such as image text editing and OCR translation~\cite{wu2019editing,zhang2019ensnet}, require to eliminate the original texts. Therefore, the fine-grained stroke-level representation can accurately define the region that needs the inpainting operation in the task.

To address aforementioned issues, 
in this work:
\begin{enumerate}
    \item Stroke Assisted Prediction Network is proposed to represent the text area from both text- and stroke-level, which is expert in detecting extremely small or low-resolution strokes. 
    \item Hierarchical Relation Graph Network is adopted to perform relational reasoning, hierarchical aggregation and linkage prediction, which are beneficial for splitting the close texts and making arbitrary-shaped text region more precisely located. The whole process is shown in Fig.\ref{new_s}.
    \item {\it SynthStroke} is introduced to promote the research in text detection. Significantly, it includes 800 thousand synthetic images with both the text- and stroke-level annotations.
Compared with the commonly used {\it SynthText}~\cite{gupta2016synthetic} in this field, {\it SynthStroke} is more challenging and promising to train a more powerful text detector.
\end{enumerate}

\section{Related Work}

\noindent 
\textbf{Regression or Segmentation based methods.} Regression based methods usually localize text boxes by directing the offsets from anchors or pixels. For instance, TextBoxes~\cite{liao2017textboxes} modified the shape of convolution kernels to effectively capture the text with various aspect ratios. LOMO~\cite{zhang2019look} tried to iteratively refine bounding box proposals. However, regression-based methods often require complex anchor setting and exhaustive tuning, and most of them are limited to represent accurate bounding boxes for arbitrary-shaped texts. Segmentation-based methods formulate text detection as a segmentation problem.
TextSnake~\cite{long2018textsnake} reconstructed the texts with the estimated geometry attributes. PSENet~\cite{wang2019shape} proposed progressive scale expansion by different scale kernels to position boundaries among close texts. However, they commonly struggle with splitting the close texts, and time-consuming post-processing~\cite{deng2018pixellink} is often involved to group pixels into text instances.

It is worth noting that previous methods such as SegLink~\cite{shi2017detecting} and CRAFT~\cite{baek2019character} which segmented or regressed each rectangular box to obtain a single character while still containing background interference.
More relevantly, Strokelets~\cite{yao2014strokelets} adopted a rule based procedure to generate strokes with multi-scale representation to show its effectiveness for text recognition.
As a contrast, our StrokeNet could completely segment the characters (strokes) from complex background by accurately predicting the corresponding segmentation maps, which is more conducive to the subsequent processing of text detection.

\noindent
\textbf{Hybrid methods.} Hybrid methods combine the idea of two mainstream ideas. They typically perform the pixel-level segmentation to seek text regions and then apply bounding box regression to make the final prediction. For example, EAST~\cite{zhou2017east} predicted offsets from pixels in each text region to perform multi-oriented regression. DRRG~\cite{zhang2020deep} proposed an innovative local graph to bridge a segmentation-based text proposal model and a deep relational reasoning graph network. Hybrid methods can inherit the advantages of both sides to further improve the detection accuracy. However, these methods have not adequately exploited the abundant information contained in the fine-grained (i.e., stroke-level) representation, tending to confuse adjacent text regions for incorrect detection. Our proposal also falls into this category, by encoding the hierarchical representation of text 
region in a segmentation based manner and reasoning the relations according to regression based box proposals, the strengths of two mainstream ideas are combined deeply to complement each other.

\section{Proposed Method}
\begin{figure}[h]
\begin{center}
\includegraphics[width=.48\textwidth]{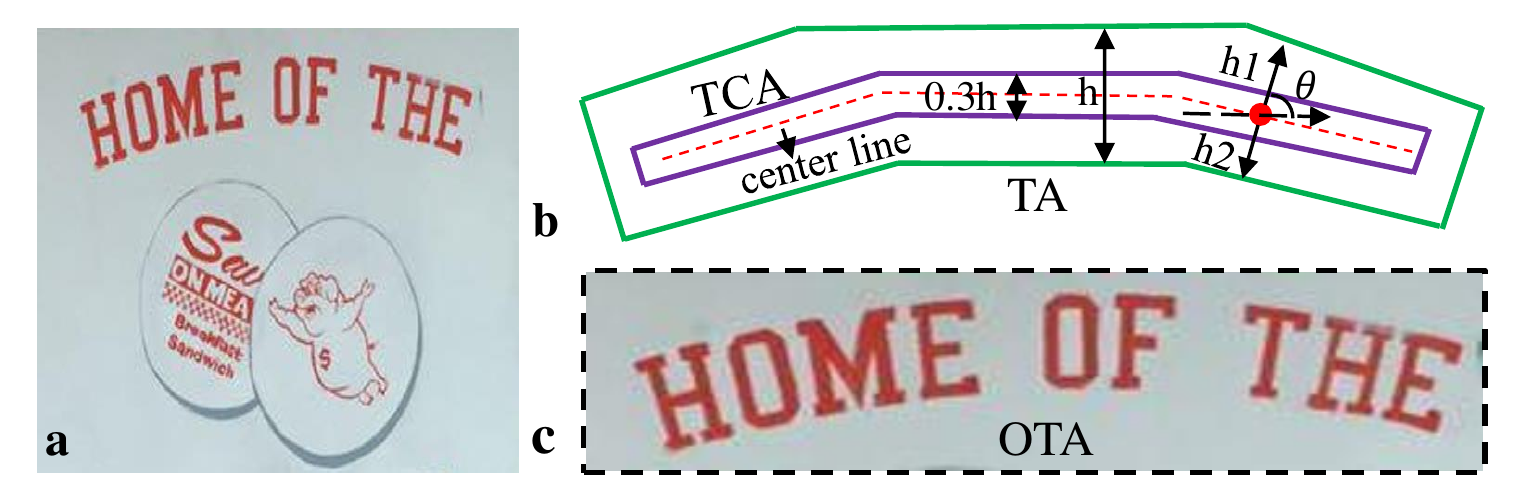}
\caption{Illustration of: (a) Original image; (b) The predicted attributes of text area; (c) The extracted outer  rectangle of TA.}
\label{new_1}
\end{center}
\end{figure}
\begin{figure*}[h]
\begin{center}
\includegraphics[width=.9\textwidth]{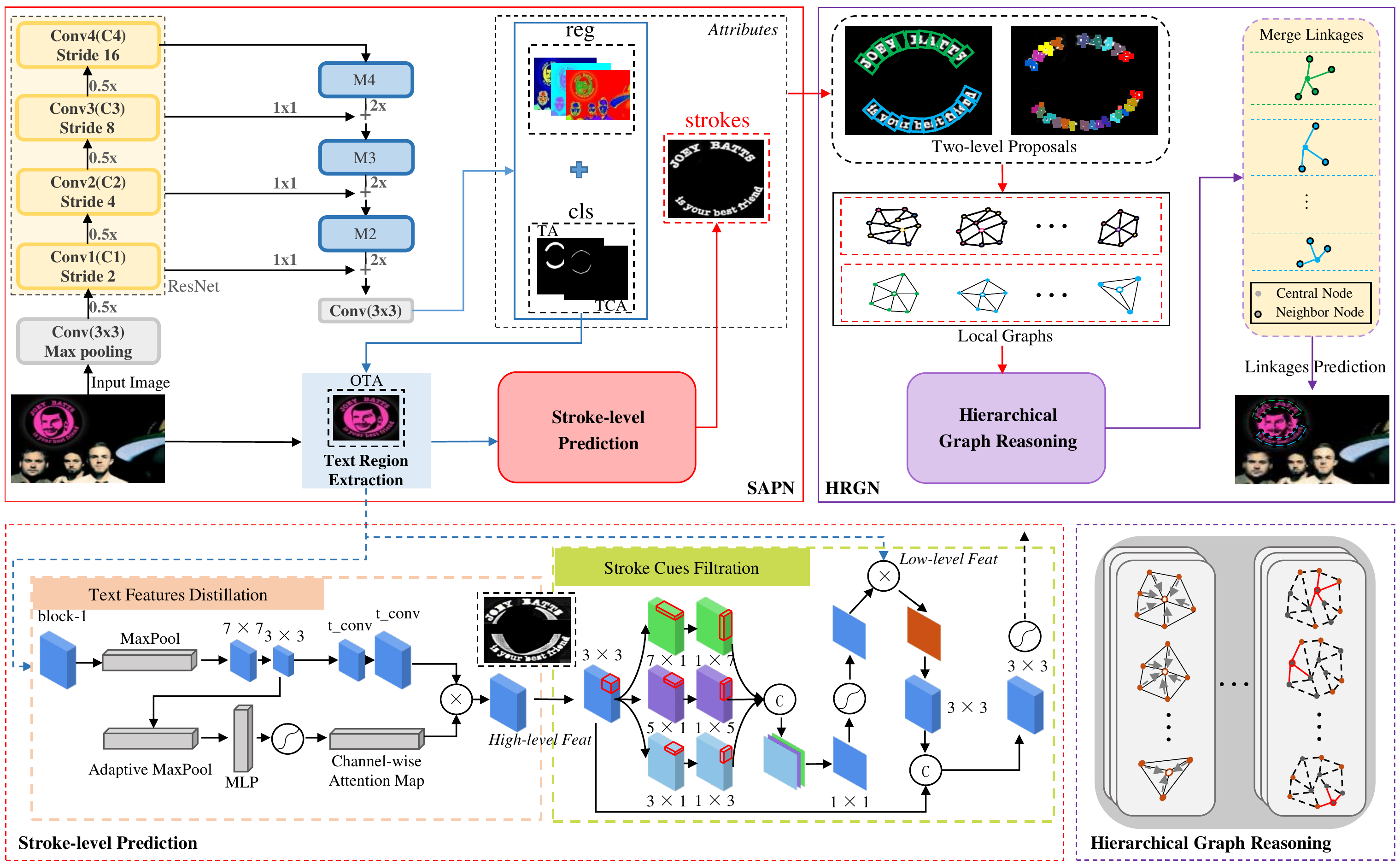}
\caption{Pipeline of StrokeNet. We detect texts accurately by locating fine-grained strokes.}
\label{fig:1}
\end{center}
\end{figure*}

The pipeline of proposed StrokeNet is illustrated in Fig.\ref{fig:1}, including two major modules namely SAPN and HRGN. In the first module, we start to apply ResNet-50
equipped with FPN~\cite{lin2017feature} as backbone, to predict the classification and regression confidence of potential text area (text-level prediction block, denoted as TLP). Then, we introduce the stroke-level prediction (SLP) block to precisely detect strokes within the predicted text area. Afterwards, box proposals of both text- and stroke- levels are extracted and treated as graph nodes to establish the corresponding local graphs. 
In the second module, the isomorphic stroke graph is first built to update the attention-guided representation among stroke-level nodes.
Then, the heterogeneous text graph is further built for relational reasoning and hierarchical aggregation from both levels.
Finally, the likelihood of linkages among text-level nodes are inferred which are grouped into holistic text instances.

\subsection{Stroke Assisted Prediction Network}
The backbone adopted in this module is conducive to preserve spatial resolution~\cite{lin2017feature} and take full advantage of high-level semantic information. After extracting the 32-channel backbone features, two consecutive convolution layers with 16 and 8 output channels are applied to predict the attributes of the text area. Concretely, 4 of the output 8 channels define the classification logits of text area (TA) and text center area (TCA), and the rest 4 channels define the regression logits of ${h\mathop{{}}\nolimits_{{1}}}$, ${h\mathop{{}}\nolimits_{{2}}}$, ${cos \theta }$, and ${sin \theta }$. As shown in Fig.\ref{new_1}, TA represents the area where the text is located, where TCA is defined by shrinking TA along the direction perpendicular to the text writing~\cite{zhang2020deep}. Besides, ${h\mathop{{}}\nolimits_{{1}}}$ and ${h\mathop{{}}\nolimits_{{2}}}$ define the distance from current pixel to the upper edge and lower edge of TA respectively. 
Furthermore, $\theta$ represents the orientation of the text, and naturally indicates an optimization constraint ${cos \theta }^2 + {sin \theta }^2 = 1$~\cite{long2018textsnake}. 
We will introduce stroke-level prediction block, where Text Features Distillation sub-block employs channel-wise attention to distill text representation from the backbone features, while Stroke Cues Filtration sub-block employs multi-scale orthogonal convolutions as well as spatial attention on rough stroke cues to suppress redundant background details.

\noindent
\textbf{Text Features Distillation.} Since the obtained pyramidal features from different layers of backbone contribute unequally to the representation of strokes, we introduce this sub-block to distill abstract semantic information of TA. Specifically, we crop out the outer rectangle of the TA (denoted by OTA) from input image, and utilize global pooling combined with consecutive convolution layers to generalize the features of OTA obtained from backbone. Then one branch up-samples the generalized features with factor 4, while the other branch adopts adaptive pooling connected by a shared MLP as well as a sigmoid layer to compute the channel-wise attention map. Finally, the two branches are multiplied to achieve semantic distillation, obtaining rough stroke cues (e.g., color, texture and edge representation of strokes) which are sent to Stroke Cues Filtration sub-block for further filtration.

\noindent
\textbf{Stroke Cues Filtration.} Generally, strokes of each text area can be regarded as the connected region surrounded by a series of edges. Inspired by previous edge detection methods~\cite{patel2007detecting,wang2020contournet}, we heuristically model fine-grained stroke-level representation from orthogonal directions. Particularly, multi-scale orthogonal convolutions are introduced to compute the attention coefficients for rough stroke cues. Besides, while the features produced by previous sub-block are incapable of providing abundant stroke details, we take the 3-channel RGB features of OTA as auxiliary stroke cues which include sophisticated texture details. Then spatial attention is performed by multiplying auxiliary stroke cues and the corresponding attention coefficients to handle rough stroke cues, performing cues filtration and removing redundant interference of background. Finally, we aggregate the outputs of two sub-blocks to produce high-quality strokes with fine-grained details.

\noindent
\textbf{Loss.} There are three losses in SAPN module, which could be formulated as:
\begin{align}
\begin{array}{l}
{L\mathop{{}}\nolimits_{{SAPN}}=L\mathop{{}}\nolimits_{{cls}}+L\mathop{{}}\nolimits_{{reg}}+L\mathop{{}}\nolimits_{{stroke}}} , \nonumber
\end{array}
\end{align}%
where ${L\mathop{{}}\nolimits_{{cls}}}$ can be further decomposed as $\lambda_1 L_{ta} + \lambda_2 L_{tca}$. 
${L\mathop{{}}\nolimits_{{ta}}}$ indicates the OHEM loss~\cite{shrivastava2016training} for TA, and ${L\mathop{{}}\nolimits_{{tca}}}$ represents the cross-entropy loss for TCA. Besides:
\begin{align}
\begin{array}{l}
{\begin{array}{*{20}{l}}{{L\mathop{{}}\nolimits_{{reg}}= \lambda \mathop{{}}\nolimits_{{3}} \left( L\mathop{{}}\nolimits_{{sin}}+L\mathop{{}}\nolimits_{{cos}} \left) +L\mathop{{}}\nolimits_{{h}}\right. \right. }} ,\nonumber \end{array}}
\end{array}
\end{align}
where ${L\mathop{{}}\nolimits_{{sin}}}$ and ${L\mathop{{}}\nolimits_{{cos}}}$ denote the regression loss for the predicted angles. 
For ${L\mathop{{}}\nolimits_{{h}}}$, we adopt the method in~\cite{zhang2020deep} to obtain the loss of height regression.

Furthermore, we employ the hybrid loss to guide stroke detection, which is defined as $L_{stroke} = \lambda_4 L_{MSE} + \lambda_5 L_{SSIM}$. 
${L\mathop{{}}\nolimits_{{MSE}}}$ is adopted to ensure the pixel-wise accuracy, while stroke structure optimization is guided via ${L\mathop{{}}\nolimits_{{SSIM}}}$~\cite{wang2004image}. During training, ${\mathop{{ \lambda }}\nolimits_{{1}}}$, ${\mathop{{ \lambda }}\nolimits_{{2}}}$, ${\mathop{{ \lambda }}\nolimits_{{3}}}$, ${\mathop{{ \lambda }}\nolimits_{{4}}}$ and ${\mathop{{ \lambda }}\nolimits_{{5}}}$ are tunable but simply set to ${1}$ for all experiments. 
The detailed equations for ${L\mathop{{}}\nolimits_{{sin}}}$, ${L\mathop{{}}\nolimits_{{cos}}}$, $L_h$, $L_{MSE}$ and $L_{SSIM}$ can refer to the supplementary materials. 

\begin{figure*}[h]
\centering
\includegraphics[width=.95\textwidth]{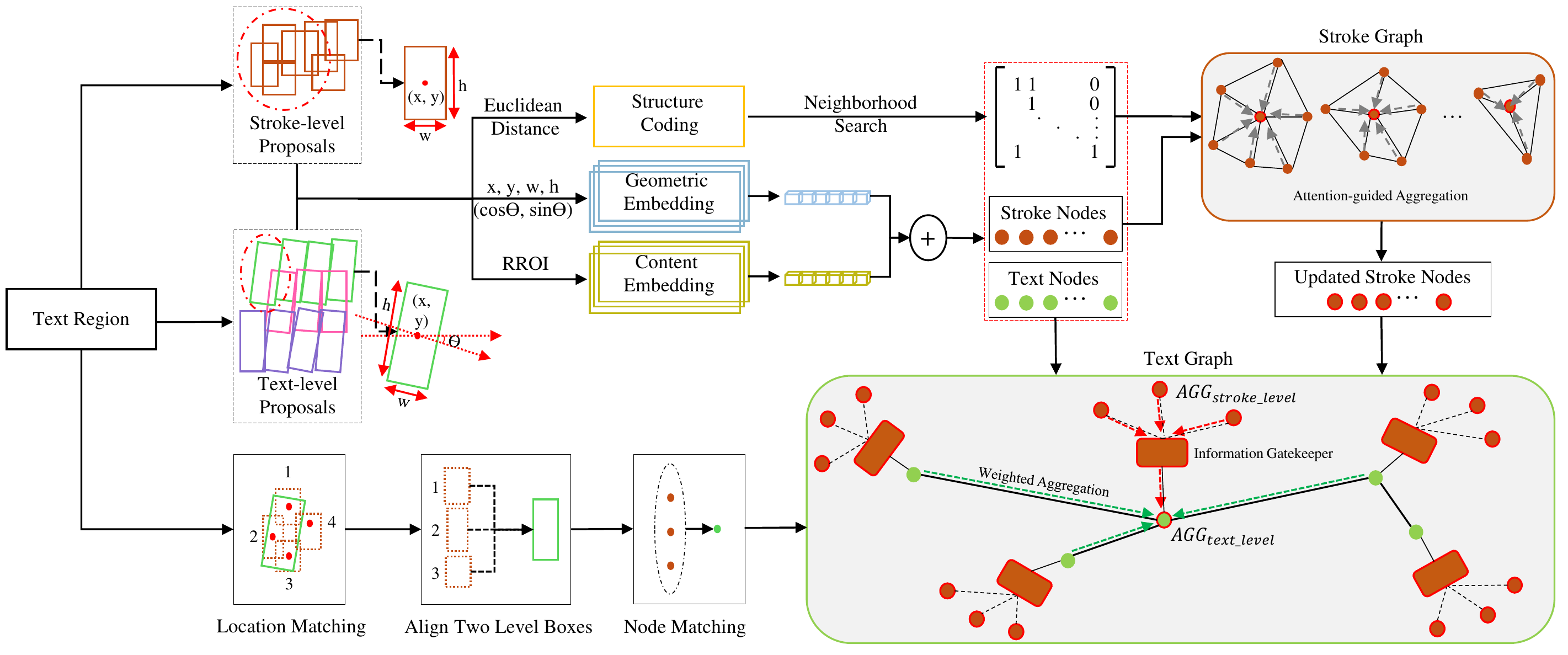}
\caption{Illustrations of graph generation and feature aggregation.}
\label{fig:2}
\end{figure*}
\subsection{Hierarchical Relation Graph Network}

Since each text instance could be divided into a series of ordered quadrilateral components along the direction of the text writing, a isomorphic stroke graph and a heterogeneous text graph are built separately for hierarchical relations reasoning and linkages prediction by extracting candidate bounding boxes from both text- and stroke- levels. As shown in Fig.\ref{fig:2}, we extract a series of text-level proposals within the predicted text area by following the method in~\cite{long2018textsnake}, while shrinking the size of boxes to obtain corresponding stroke-level proposals within the detected region of strokes. In the meantime, NMS~\cite{rothe2014non} and boundary determination are introduced to limit the total number of generated proposals (graph nodes). Please refer to the supplementary materials of graph generation for more details.

We adopt complementary representation for feature initialization of nodes at both levels, i.e., geometric embedding and content embedding. Concretely, circular functions~\cite{vaswani2017attention} are applied to get geometric embedding by encoding the geometric attributes into high dimensional spaces, while content embedding is obtained by sending the predicted feature map with the geometric attributes of each proposal to the RRoI-Align layer~\cite{he2017mask}. 

Based on the built isomorphic stroke graph which contains only stroke-level nodes, we first adopt the attention mechanism proposed in~\cite{velivckovic2017graph} to model diverse relationships from both aspects of structure and content for attention-guided representation.
Concretely, for a stroke node ${s}$ and its neighbor ${n}$ ${ \left( n \in N \right) }$ where ${N}$ denotes the neighbor set, an attention coefficient between them can be formulated as:
\begin{align}
\alpha_{sn} = \frac{{exp \left( LeakyReLU \left( \mathop{{a}}\nolimits^{{T}} \left[ W\mathop{{h}}\nolimits_{{s}} \oplus W\mathop{{h}}\nolimits_{{n}} \left]  \left)  \right) \right. \right. \right. \right. }}{{\mathop{ \sum }\limits_{{k \in N}}{exp \left( LeakyReLU \left( \mathop{{a}}\nolimits^{{T}} \left[ W\mathop{{h}}\nolimits_{{s}} \oplus W\mathop{{h}}\nolimits_{{k}} \left]  \left)  \right) \right. \right. \right. \right.  }}} , \nonumber
\end{align}%
where ${\mathop{{h}}\nolimits_{{s}}}$ and ${\mathop{{h}}\nolimits_{{n}}}$ denote the feature vectors of two nodes, ${W}$ and ${a}$ are trainable parameters, and $\oplus$ means concatenation.
After that, the representation of node $s$ will be updated as ${h_s}^{updated}=\text{Sigmoid}\left(\sum_{k \in N} \alpha_{sk}\cdot  Wh_{k}\right)$. 
By learning to increase the attention weight of adjacent nodes that are jointly appear in the direction along the writing, while suppressing the weight of adjacent nodes appear in the direction perpendicular to the writing or in other directions, the built stroke graph performs a distinguishable aggregation to identify the text instances that are close to each other.
The updated stroke nodes, together with text-level nodes, are then adopted for hierarchical relations reasoning in the heterogeneous text graph.

In the built text graph, each text node is connected with extra stroke-level nodes apart from its text-level neighbors. 
For a text node ${t}$, we filter out the top three nearest stroke nodes based on the distance of their centers to ${t}$. A two-stage information aggregation process is then utilized for the update of text-level representation.
In the first stage, a weighted average aggregator is employed where the weights come from the normalized adjacency matrix ${A}$ among text-level nodes, which is defined as:
\begin{equation}
    AGG_{text\_level}:\text{  } h_t^{stage_{1}} = \sum_{\forall m\in  \mathcal{N} (t)}^{} a_{t,m}h_m, \nonumber
\end{equation}
where $\mathcal{N} (t)$ indicates the $1$-hop neighbor set of $t$. 

Considering that diverse stroke nodes contain information from different parts of the text area, contributing distinctly to the representation of each text node. 
In the second stage, we perform a expressive information aggregation step from stroke nodes to text nodes. 
In detail, a soft mask is first computed as the following:
\begin{gather}
    \tilde{h_t} = Meanpooling(\mathcal{F} (\tilde{\mathcal{N}}(t))), \nonumber \\ 
    m(h_t)=\text{Sigmoid} (MaxPooling(\mathcal{F} \tilde{(\mathcal{N}}(t)) \cdot M\cdot \tilde{h_t})), \nonumber
\end{gather}
where $\tilde{\mathcal{N}}(t)$ indicates the stroke-level $1$-hop neighbor set of $t$, while $\mathcal{F}(.)$ denotes the corresponding feature vectors. ${M}$ is a trainable weight matrix, and “$.$" represents the matrix multiplication. 
The obtained $m(h_t)$ serves as the information gatekeeper, which will be multiplied by the feature of stroke-level neighbors in the heterogeneous graph:
\begin{equation*}
    AGG_{stroke\_level}:\text{  } h_t^{stage_{2}} = \mathcal{F} \tilde{(\mathcal{N}}(t))\otimes m(h_t), 
\end{equation*}
where $ \otimes $ denotes the element-wise product. 
In this way, the stroke-level aggregation for each text node is restricted to a dynamic sub-part of the whole graph, and the informative stroke nodes will be encouraged to perform aggregation operations and the leftovers will be penalized. 
Besides, this mechanism is conductive to eliminating irrelevant nodes when learning local details, resulting in an efficient learning architecture while stabilizing the training process.

After that, the aggregation from both levels are fused by a gated sum function:
\begin{equation*}
    h_t^{updated} = Fuse(AGG_{text\_level}, AGG_{stroke\_level}), 
\end{equation*}
where $Fuse(a, b)=p\cdot a+(1-p)\cdot b$, and $p=\text{Sigmoid} (W_p[a;a\otimes b;b]+b_p)$. $W_p$ and $b_p$ are trainable parameters.
Finally, all updated representation of text nodes (denoted as $H$) are utilized to predict the linkage relations from each center node to its neighbors, by the modified graph convolution~\cite{zhang2020deep}: $P=\text{Softmax} ((H\oplus L\cdot H)W_p)$,
%
%
where ${L}$ denotes symmetric normalized Laplacian of the adjacency matrix ${A}$, and $W_p$ is the weight parameter. 
The outputs from the last graph layer are used to predict linkages which are finally grouped for locating arbitrary-shaped text instances. 
The cross-entropy loss is adopted for training. 

During inference, we first apply Stroke Assisted Prediction Network to obtain the multi-level predictions of each text instance, which are thresholded for constructing multiple local graphs. 
Next, Hierarchical Relation Graph Network is introduced to infer the relations at both levels and make linkages prediction among text-level nodes. 
According to the classification results, text nodes are grouped by Breath First Search method and sorted by Min-Path algorithm, to obtain the boundary of arbitrary-shaped text by sequentially linking the mid-point of both the top and the bottom in ordered text nodes.

\begin{table*}[t]
\centering
\scalebox{0.9}{
\begin{tabular}{l|c|c|c|c|c|c|c|c|c}
\toprule
\multicolumn{1}{c|}{\multirow{2}{*}{\textbf{Method}}} & \multicolumn{3}{c|}{\textbf{CTW-1500}}                 & \multicolumn{3}{c|}{\textbf{Total-Text}}               & \multicolumn{3}{c}{\textbf{ICDAR 2015}}                           \\ \cline{2-10} 
                         & \textbf{Recall}             & \textbf{Precision}             & \textbf{Hmean}             & \textbf{Recall}             & \textbf{Precision}             & \textbf{Hmean}             & \textbf{Recall}             & \textbf{Precision}             & \textbf{Hmean}    \\ \toprule
TextSnake~\cite{long2018textsnake}                    & 85.3                                 & 67.9                                    & 75.6         & 74.5                                 & 82.7                                    & 78.4          & 84.9            & 80.4               & 82.6 \\ \hline
PSENet~\cite{wang2019shape}                   & 79.7          & 84.8          & 82.2          & 84.0          & 78.0          & 80.9          & 84.5          & 86.9          & 85.7 \\ \hline
CRAFT~\cite{baek2019character}                    & 81.1          & 86.0          & 83.5          & 79.9          & 87.6          & 83.6          & 84.3          & 89.8          & 86.9 \\ \hline
DB~\cite{liao2020real}                       & 80.2          & 86.9          & 83.4          & 82.5          & 87.1          & 84.7          & 83.2          & 91.8          & 87.3 \\ \hline
ReLaText~\cite{ma2021relatext}                  & 83.3          & 86.2          & 84.8          & 83.1          & 84.8          & 84.0          & -             & -             & -                         \\ \hline
DRRG~\cite{zhang2020deep}                 & 83.0          & 85.9          & 84.5          & 84.9          & 86.5          & 85.7          & 84.7          & 88.5          & 86.6                      \\ \hline
ContourNet~\cite{wang2020contournet}                & 84.1          & 83.7          & 83.9          & 83.9          & 86.9          & 85.4          & 86.1          & 87.6          & 86.9                      \\ \hline
ABCNet~\cite{liu2020abcnet}                   & 78.5          & 84.4          & 81.6          & 81.3          & 87.9          & 84.5          & -             & -             & -                         \\ \hline
FCENet~\cite{zhu2021fourier}                   & 83.4          & 87.6          & 85.5          & 82.5          & 89.3          & 85.8          & 82.6          & 90.1          & 86.2                      \\ \hline
SDM-ResNet-50~\cite{yaosequential}                  & 84.4                                 & \textbf{88.4}                                    & 86.4           & 86.0                                 & \textbf{90.1}                                    & 88.4          & 89.2            & \textbf{92.0}               & \textbf{90.6}                         \\\toprule
StrokeNet (T)       &\textbf{86.3}  &88.2  &\textbf{87.2} &\textbf{87.8}  &89.0  &\textbf{88.4}  &\textbf{89.2}  &91.7  &90.4              \\ \hline
\textbf{StrokeNet (S)}        &\textbf{86.9}  &\textbf{88.7}  &\textbf{87.8}  &\textbf{88.2}  &\textbf{89.5}  &\textbf{88.8}  &\textbf{89.6}  &\textbf{92.3}  &\textbf{90.9}             \\ \bottomrule   
\end{tabular}
}
\caption{Experimental results on CTW-1500, Total-Text and ICDAR 2015. The top two best scores are highlighted in bold.}
\label{tab-total}
\end{table*}

\begin{table*}[t]
\centering
\scalebox{0.9}{
\begin{tabular}{l|c|c|c|l|c|c|c|l|c|c|c}
\toprule
\multicolumn{1}{c|}{\multirow{2}{*}{\textbf{Method}}} & \multicolumn{3}{c|}{\textbf{MSRA-TD500}}                     & \multicolumn{1}{|c|}{\multirow{2}{*}{\textbf{Method}}} & \multicolumn{3}{c|}{\textbf{ICDAR 2017 MLT}}  & \multicolumn{1}{|c|}{\multirow{2}{*}{\textbf{Method}}} & \multicolumn{3}{c}{\textbf{ICDAR 2019 MLT}}      \\ \cline{2-4} \cline{6-8} \cline{10-12} 
                                  & \textbf{R}    & \textbf{P}    & \textbf{H}     &                                   & \textbf{R}    & \textbf{P}    & \textbf{H}    &                          & \textbf{R}    & \textbf{P}    & \textbf{H}    \\ \toprule
EAST~\cite{zhou2017east}                & 61.6            & 81.7               & 70.2   & He et al.~\cite{he2018multi}          & 57.9            & 76.7               & 66.0    & CLTDR~\cite{nayef2019icdar2019}                & 54.0            & 77.2               & 63.5 \\ \hline
SegLink~\cite{shi2017detecting}                           & 70.0          & 86.0          & 77.0           & Lyu et al.~\cite{lyu2018multi}         & 55.6            & 83.8      & 66.8          & PSENet~\cite{wang2019shape}                   & 59.6          & 73.5          & 65.8          \\ \hline
PixelLink~\cite{deng2018pixellink}                         & 73.2          & 83.0          & 77.8                    & DRRG~\cite{zhang2020deep}          & 61.0            & 75.0               & 67.3          & RRPN~\cite{ma2018arbitrary}                     & 63.0          & 77.7          & 69.6          \\ \hline
TextSnake~\cite{long2018textsnake}                         & 73.9          & 83.2          & 78.3                    & LOMO~\cite{zhang2019look}               & 60.6            & 78.8               & 68.5          & CRAFT~\cite{baek2019character}                    & 62.7          & 81.4          & 70.9          \\ \hline
CRAFT~\cite{baek2019character}                             & 78.2          & 88.2          & 82.9                   & CRNet~\cite{zhou2020crnet}               & 64.1            & 84.3               & 72.8          &  MaskRCNN++~\cite{nayef2019icdar2019}                     & 78.2          & 82.6          & 80.4          \\ \hline
PAN~\cite{wang2019efficient}                 & 83.8            & 84.4               & 84.1    &DB~\cite{liao2020real}         & 67.9            & 83.1      & 74.7    &PMTD~\cite{liu2019pyramid}                     & 78.1          & 87.5          & 82.5 \\ \hline
DRRG~\cite{zhang2020deep}          & 82.3            & 88.1               & 85.1    &SBD~\cite{liu2019omnidirectional}               & 70.1            & 83.6               & 76.3    & Multi-stage~\cite{nayef2019icdar2019}          & 79.8            & \textbf{87.8}               & 83.6 \\ \hline
ReLaText~\cite{ma2021relatext}                          & 83.2          & \textbf{90.5} & \textbf{86.7}                   & SDM~\cite{yaosequential}                                & 75.3          & \textbf{86.8} & 80.6            & Tencent-DPPR~\cite{nayef2019icdar2019}       & 80.1          & 87.5          & 83.6          \\ \toprule
StrokeNet (T)                &\textbf{85.2}  &87.6           & 86.4         & StrokeNet (T)            &\textbf{77.2}  &85.8           &\textbf{81.3}  & StrokeNet (T)   & \textbf{84.7} &87.3  &\textbf{86.0}  \\ \hline
\textbf{StrokeNet (S) }                & \textbf{85.6} &\textbf{88.3}           &\textbf{86.9}        & \textbf{StrokeNet (S) }           &\textbf{78.4}  &\textbf{87.1}           &\textbf{82.5} & \textbf{StrokeNet (S) }  &\textbf{86.5}  &\textbf{88.4}           &\textbf{87.4} \\ 
\bottomrule
\end{tabular}
}
\caption{Experimental results on MSRA-TD500, ICDAR 2017 MLT and ICDAR 2019 MLT benchmarks. The performance of comparative methods on ICDAR 2019 MLT are reported in~\cite{nayef2019icdar2019}. R: Recall, P: Precision, H: Hmean. The top two best scores are highlighted in bold.}
\label{tab-left}
\end{table*}

\begin{table*}[t]
\centering
\scalebox{0.9}{
\begin{tabular}{l|c|c|c|c|c|c}
\toprule
\multicolumn{1}{c|}{\multirow{2}{*}{\textbf{Method}}} & \multicolumn{6}{c}{\textbf{FPS}}                                                                        \\ \cline{2-7} 
\multicolumn{1}{c|}{}                        & \textbf{CTW-1500} & \textbf{Total-Text} & \textbf{ICDAR 2015} & \textbf{MSRA-TD500} & \textbf{ICDAR 2017 MLT}        & \textbf{ICDAR 2019 MLT}        \\ \toprule
PSENet~\cite{wang2019shape}                                        & 3.9      & 3.9        & 1.6        & -          &      -                 &  -                     \\ \hline
ContourNet~\cite{wang2020contournet}                                    & 4.5      & 3.8        & 3.5        & -          &   -                    &  -                     \\ \hline
PAN~\cite{wang2019efficient}                                            & 39.8     & 39.6       & 26.1       & -          &   -                    &  -                     \\ \hline
DB (ResNet-50)~\cite{liao2020real}                                & 22.0     & 32.0       & 12.0       & 32.0       &    19                   &    -                   \\ \hline
TextFuseNet~\cite{ye2020textfusenet}                      & 3.7      & 3.3        & 4.1        & -          &    -                   &      -                 \\ \hline
ReLaText~\cite{ma2021relatext}                                      & 10.6     & -          & -          & 8.3        &     -                  &    -                   \\ \hline
SAE(720)~\cite{tian2019learning}                                      & 3.0      &   -         & 3.0        &   -         &    -                   &   -                    \\ \toprule
\textbf{StrokeNet (T/S)}                               &  6.0        & 6.0          & 5.2           & 5.9           & 5.5 & 5.6 \\ \bottomrule
\end{tabular}
}
\caption{Compare the detection speed of different methods.}
\label{speed}
\end{table*}

\section{Experiments}
\subsection{Benchmarks and Implementation Details}
\begin{figure}
\begin{center}
\includegraphics[width=.49\textwidth]{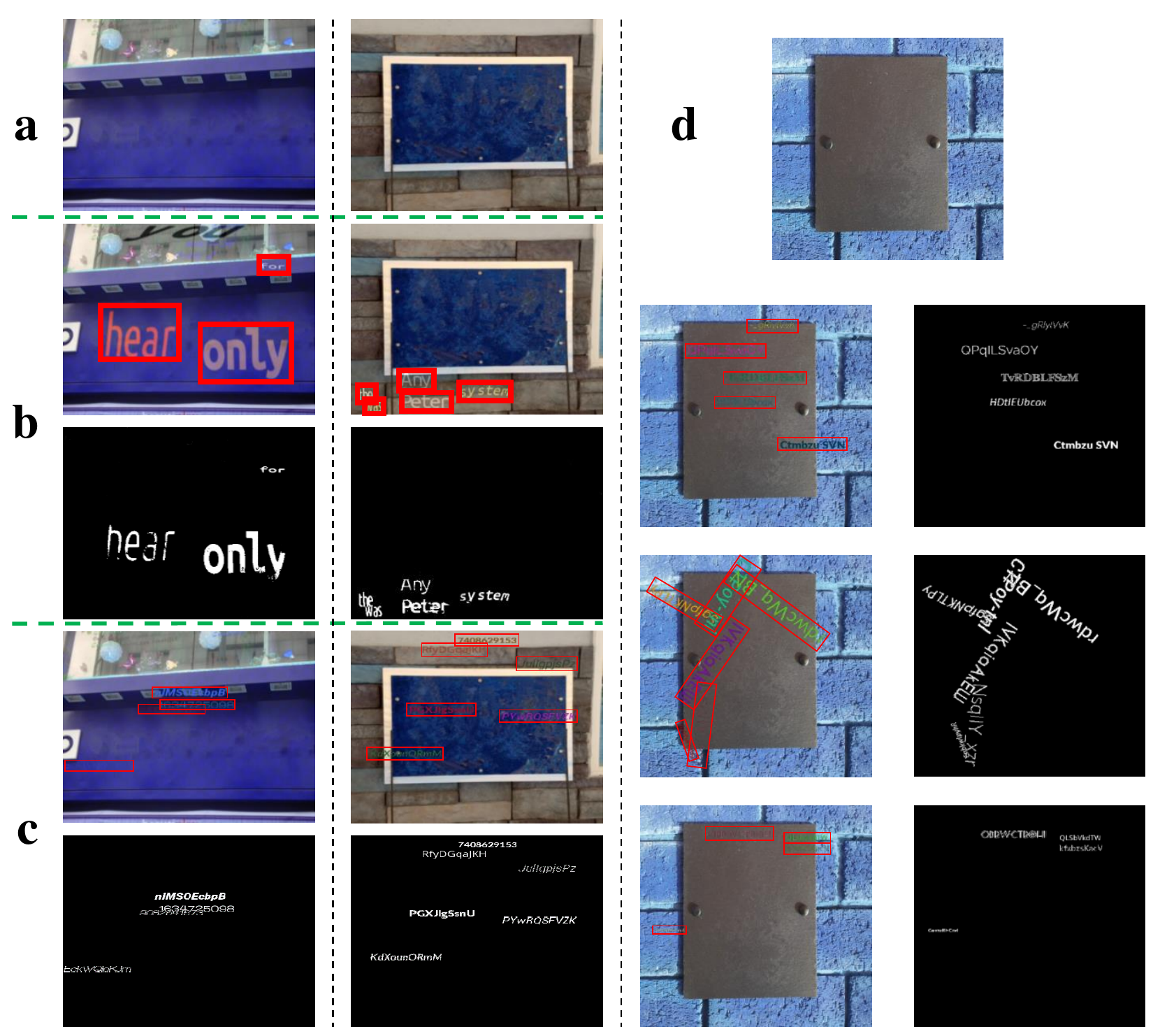}
\caption{Illustration of: (a) Original images; (b) The synthesized images in {\it SynthText} and the pseudo stroke labels we extracted. (c) The synthesized images in our {\it SynthStroke} with stroke-level annotations. (d) {\it SynthStroke} contains many characteristics of text such as blending with the background, rotating at any angle, and extremely small presentation, etc.}
\label{new_5}
\end{center}
\end{figure}

We evaluate StrokeNet on six standard benchmarks: CTW-1500, Total-Text, MSRA-TD500, ICDAR2015, ICDAR 2017 MLT and ICDAR 2019 MLT. 
Besides, we introduce {\it SynthStroke} 
to pre-train the whole framework, which is helpful for subsequent performing fine-tuning and evaluation on real scene benchmarks. 
We synthesize the whole dataset based on the 8000 native images, which contain no texts and are collected from the open public repository \footnote{https://github.com/HCIILAB/Scene-Text-Removal}. 
Specifically, {\it SynthStroke} consists of 800 thousand synthetic images with approximately 8 million synthetic word instances.
It is noted that another synthetic dataset, namely {\it SynthText}~\cite{gupta2016synthetic}, which is also synthesized from the mentioned resource and commonly applied for the pre-training of many text detectors in previous research.  
A visual comparison of our {\it SynthStroke} and {\it SynthText} is summarized in Fig.\ref{new_5}, and please refer to the supplements for details.

For the whole framework, we first pre-train our StrokeNet with the introduced {\it SynthStroke} for 5 epochs, and then perform fine-tuning on benchmark datasets for 1000 epochs by extracting their pseudo stroke labels. The model obtained in this way is denoted as \textbf{StrokeNet (S)}.
To improve fairness, we pre-train another model namely \textbf{StrokeNet (T)} on {\it SynthText} after extracting its pseudo stroke labels, and use the same evaluation criteria on benchmarks.
All experiments are performed on a single image resolution. A detailed descriptions of benchmarks, datasets and implementations could be found in the supplementary materials.

\subsection{Comparison with State-of-the-Art Methods}

\noindent
\textbf{Close and arbitrary-shaped text detection.} We compare StrokeNet with several state-of-the-art methods on two curved benchmarks in Table \ref{tab-total}, including CTW-1500 and Total-Text. 
Benefiting from the introduced HRGN, our method achieves promising results on representing close and arbitrary-shaped texts especially with varying degrees of curvature.

\noindent
\textbf{Small and low-resolution text detection.} We evaluate our method on ICDAR 2015, which contains a lot of small and low-resolution text instances. 
As shown in Table \ref{tab-total}, our StrokeNet achieves consistent and competitive performance in recall, precision and H-mean, because the introduced SAPN module plays an important role in effectively capturing the representation of small and low-resolution strokes. 

\begin{figure*}
	\centering
	\includegraphics[width=.7\textwidth]{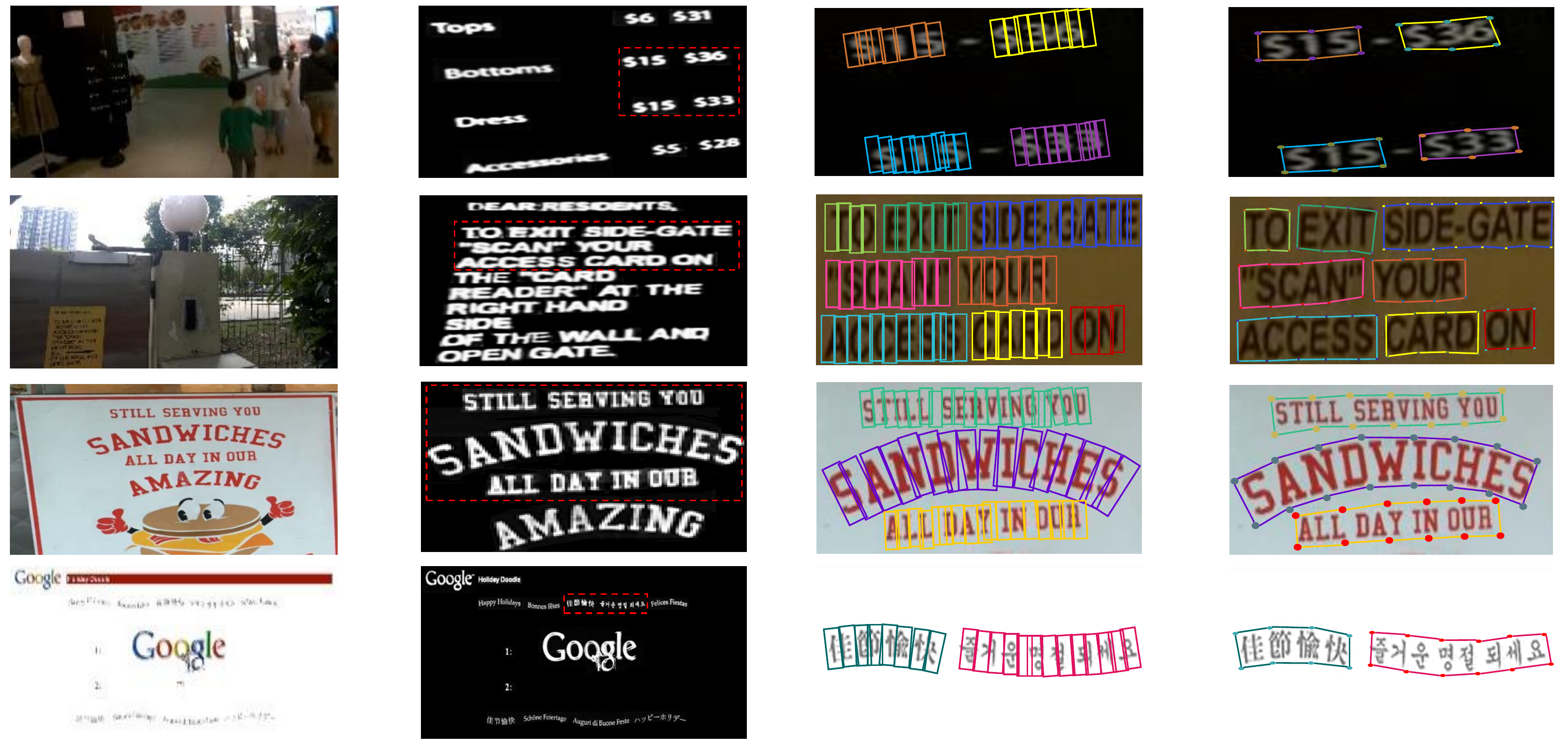}
	\caption{Visualizations on benchmarks. First column: input images, Second column: predicted strokes (only sample part of each whole area by red dotted rectangles for subsequent display). Third column: text-level boxes generation, Fourth column: detected texts.}
	\label{img4}
\end{figure*}

\noindent
\textbf{Multi-language text detection.} To test the robustness of StrokeNet to multiple languages with long texts, we evaluate our method on MSRA-TD500, ICDAR 2017 MLT and ICDAR 2019 MLT benchmarks. The quantitative results are listed in Table \ref{tab-left}.
Significantly, the evaluations on ICDAR 2019 MLT benchmark verify that StrokeNet achieves superior performance with continuous stability on large-scale dataset.
Qualitative results shown in Fig.\ref{img4} can demonstrate the effectiveness of proposed method in above three aspects.

\begin{figure*}
	\centering
	\includegraphics[width=.95\textwidth]{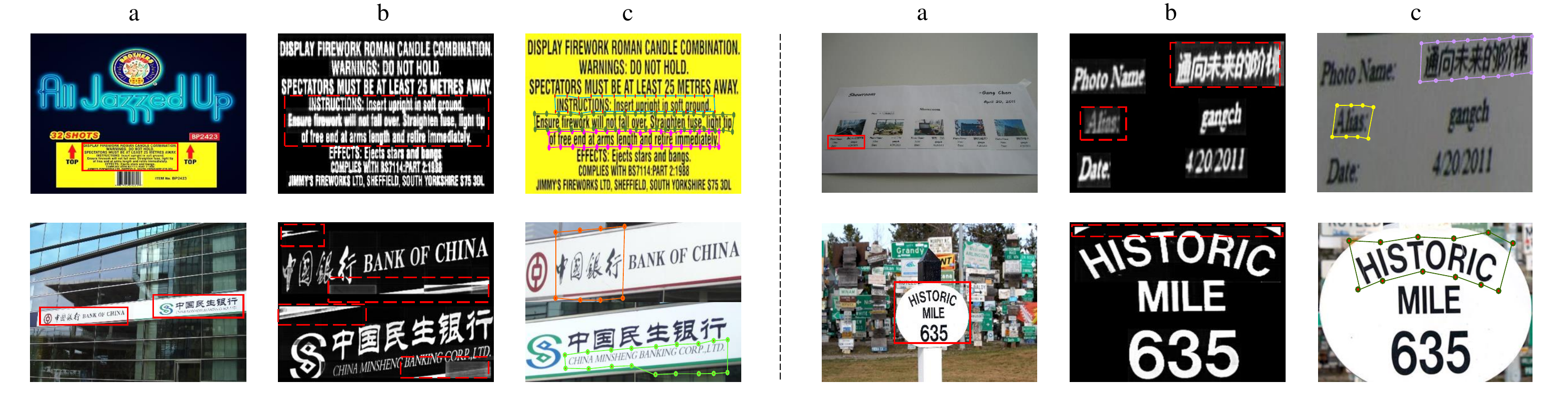}
	\caption{Some failure cases. Column a: input images with the solid red rectangles as the detection focus. Column b: the corresponding stroke-level outputs, where the dotted rectangular boxes indicate the detection problems. Column c: the final text detection results.}
	\label{img5}
\end{figure*}

Besides, the comparison of detection speed is provided in Table \ref{speed}.
As a multi-task model and the graph networks introduced in our proposal supposedly require more inference time, the speed of StrokeNet is acceptable.
Furthermore, Fig.\ref{img5} indicates some of the failure cases produced by StrokeNet. The examples shown in the first row suggest that few strokes detected by our method are blurred, but this problem does not tend to have a distinct impact on the detection of the corresponding text area.
The second row shows that StrokeNet may mistakenly detect the edges of some background objects in stroke-level detection, leading to the detection results containing undesirable background. This problem is mainly due to the introduction of orthogonal convolutions, which are sensitive to the edge features. In the future, we consider further optimization on detecting more accurate strokes while suppressing background edges.

\begin{table}[t]
	\centering
    \setlength{\tabcolsep}{1.5pt}
	\begin{tabular}{c|l|l|c|c|c|c}
		\toprule
		 \textbf{Model}      & \textbf{Module\_1} & \textbf{Module\_2} & \textbf{R} & \textbf{P} & \textbf{H} & \textbf{Gain(\%)} \\ \toprule
		 baseline (S)   & TLP                   & NONE                    & 71.6            & 73.8               & 72.7        & ——   \\  \hline
		 Variant\_1 (S) & TLP+SLP               & NONE                   & 74.2            & 79.9               & 77.0        & 5.9  \\  
		 Variant\_2 (S)  & TLP                   & TG$^*$                    & 81.5            & 83.9               & 82.7        & 13.8   \\ \hline
		 StrokeNet (T)  & TLP+SLP               & SG+TG           & 85.2   & 87.6      & 86.4 & 18.8 \\
		 \textbf{StrokeNet (S)}  & TLP+SLP               & SG+TG           & \textbf{85.6}   & \textbf{88.3}      & \textbf{86.9} & \textbf{19.5} \\ \bottomrule
	\end{tabular}
	\caption{Ablation study for the main modules of our StrokeNet on MSRA-TD500 benchmark.}
	\label{tab5}
\end{table}

\begin{table}
\centering
\setlength{\tabcolsep}{1.5pt}
\begin{tabular}{c|c|c|c|c|c|c|c|c}
\toprule
\multirow{2}{*}{$L_{cls}$} & \multirow{2}{*}{$L_{reg}$} & \multirow{2}{*}{$L_{stroke}$} & \multicolumn{3}{c|}{\textbf{ICDAR 2015}} & \multicolumn{3}{c}{\textbf{ICDAR 2017 MLT}} \\ \cline{4-9} 
                      &                       &                          & \textbf{R}         & \textbf{P}        & \textbf{H}        & \textbf{R}          & \textbf{P}          & \textbf{H}         \\ \toprule
\Checkmark                     & \Checkmark                     & \XSolid                        &  84.2         &  86.1        &  85.1        &  65.6          & 75.9           &  70.4         \\ \hline
\XSolid                     & \Checkmark                     & \Checkmark                        &  87.2         &   88.7       &  87.9        & 75.0           &  83.7          &   79.1        \\ \hline
\XSolid                     & \XSolid                     & \Checkmark                       &  69.4         & 65.8         &  67.6        &  49.6          & 45.7           &  47.6         \\ \hline
\Checkmark                     & \Checkmark                     & \Checkmark                        & \textbf{89.6}           & \textbf{91.7}         & \textbf{90.6}        &  \textbf{78.4}          &  \textbf{87.1}          &  \textbf{82.5}         \\ \toprule
\end{tabular}
\caption{Ablation study of each loss term in SAPN module.}
\label{loss_ablation_study}
\end{table}

\begin{table}[h]	
	\centering
	\setlength{\tabcolsep}{1.5pt}
	\scalebox{0.9}{
	\begin{tabular}{l|l|c|c|c|c|c}
		\toprule
		\textbf{Benchmark}    & \textbf{Method}   & \textbf{Pre-train}        & \textbf{R} & \textbf{P} & \textbf{H} & \textbf{Gain(\%)} \\ \bottomrule
		CTW-1500  &EAST~\cite{zhou2017east}   & {\it SynthText}       & 49.1            & 78.7               & 60.4      & ——     \\ 
		  &\cite{zhou2017east}    & {\it SynthStroke}   & \textbf{52.3}            & \textbf{80.2}      & \textbf{63.3}  & \textbf{4.8}        \\\hline
		ICDAR2015  &DRRG~\cite{zhang2020deep}       & {\it SynthText}   & 84.7            & 88.5               & 86.6       & ——    \\ 
		 &\cite{zhang2020deep}           & {\it SynthStroke}    &  \textbf{86.4}          & \textbf{88.9}               &   \textbf{87.6}    & \textbf{1.2}     \\ 
		 &\cite{zhang2020deep} + SLP   & {\it SynthStroke}     &  \textbf{87.2}           &  \textbf{89.2}     & \textbf{88.2}     &  \textbf{1.8}     \\ \toprule
		  ICDAR2017  &DRRG~\cite{zhang2020deep}     & {\it SynthText}      & 61.0            & 75.0               & 67.3        & ——   \\ 
		 MLT &\cite{zhang2020deep}      & {\it SynthStroke}          & \textbf{64.4}           &     \textbf{75.3}           & \textbf{69.4} &  \textbf{3.2}         \\
		 &\cite{zhang2020deep} + SLP      & {\it SynthStroke}          & \textbf{65.5}           &     \textbf{76.1}           & \textbf{70.4} &  \textbf{4.6}         \\ \bottomrule
	\end{tabular}
	}
	\caption{Ablation study of introduced {\it SynthStroke} and the SLP block. The modified model (DRRG+SLP) is first pre-trained on our {\it SynthStroke}, and then evaluated on real scene benchmarks.}
	\label{sp_ablation_study_1}
\end{table}


\subsection{Ablation Study}
For our proposal, the first module SAPN contains two blocks, marked as text-level prediction (TLP) block and stroke-level prediction (SLP) block. While the second module HRGN includes two main parts, including the built stroke graph (SG) and text graph (TG). 
We conduct ablation study by removing SP and SG, leading to the variant (TLP + TG$^*$) which only adopts text-level outputs for subsequent single-level graph reasoning process. TG$^*$ means that there is no stroke nodes for the built text graph, and only text-level aggregation is performed. 
Table \ref{tab5} summarizes the results of our models with different settings on MSRA-TD500. We adopt TLP block as our baseline, which performs text-level prediction and only obtains the rectangle result of each text instance. Then we introduce the SLP block to model fine-grained stroke-level representation in a way of multi-task learning, improving the performance by 5.9\% in Hmean. 
When TG$^*$ is further added, single-level relations reasoning and aggregation are performed to localize arbitrary-shaped texts. 
Finally, we further introduce SG to conduct attention-guided aggregation and the feature fusion between nodes of both levels, greatly promoting the final results to 86.9\% Hmean rate. 

We further explore the effectiveness of each loss term in SAPN module and evaluate on ICDAR 2015 and ICDAR 2017 MLT benchmarks. As shown in Table \ref{loss_ablation_study}, the combination of $L_{reg}$ and $L_{stroke}$ obtains higher evaluation metrics than the association of $L_{cls}$ and $L_{stroke}$, indicating that the use of fine-grained segmentation loss for stroke-level prediction is more effective than adopting the coarse-grained classification loss.
Besides, although the introduced $L_{stroke}$ is essential for performance promotion, but if only adopted, limited performance will be obtained.

\subsection{Justification of External Dataset}
We perform a justification of introduced dataset by showing a significant improvement of StrokeNet after pre-training on {\it SynthStroke} compared to {\it SynthText} in Table \ref{tab-total}, \ref{tab-left} and \ref{tab5}.
Besides, we apply our {\it SynthStroke} to pre-train comparative methods and the results are shown in Table~\ref{sp_ablation_study_1}.
From the table, it is concluded that {\it SynthStroke} is beneficial for off-line pre-training of widely methods.
Moreover, we modify DRRG~\cite{zhang2020deep} which is a recently proposed text detector that also adopts the graph networks, by adding the SLP block on top of its text region detection module. In the meantime, training is performed in a way of multi-task learning, while both locations and stroke-level masks are output during inference. The corresponding results are revealed in Table~\ref{sp_ablation_study_1}, demonstrating the applicability of stroke-level representation for related methods in text detection field.
It also shows that the exploration in stroke-level representation of text area is essential for further promotion of performance.

\begin{figure}[t]
\centering
\includegraphics[width=0.8\columnwidth]{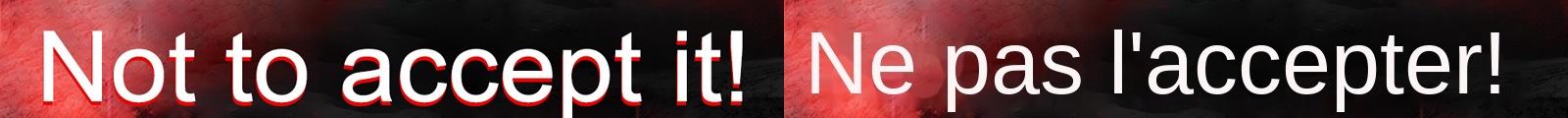}
\caption{An example of OCR translation (English to French) with StrokeNet. Please refer to supplements for more details.}
\label{fig:ocr_translate}
\end{figure}

\noindent
\textbf{Application} We have developed an OCR translation tool based on StrokeNet, shown in Fig.\ref{fig:ocr_translate}. 

\section{Conclusion}

We propose StrokeNet accompanied by {\it SynthStroke} dataset with stroke-level annotations for scene text detection. 
Our method focuses on multi-level representations and hierarchical relations reasoning of text regions, efficiently detecting extremely small or low-resolution strokes and effectively splitting close or arbitrary-shaped texts. 
For future work, it is worth to explore the end-to-end fashion of text detection to support various downstream tasks, such as text recognition and text removal.

\newpage
{\small
\bibliographystyle{ieee_fullname}
\bibliography{ms}
}

\newpage
\appendix
\section{Datasets}

\subsection{Benchmark Datasets}

We perform fine-tuning of our StrokeNet on real scene datasets, by extracting the pseudo stroke labels of several popular benchmark datasets. 
Specifically, we adopt our {\it SynthStroke} to pre-train the introduced stroke-level prediction block (SLP), and then we use the pre-trained SLP to detect the rough strokes of each text region in benchmark datasets, including CTW-1500, Total-Text, MSRA-TD500, ICDAR 2015, ICDAR 2017 MLT and ICDAR 2019 MLT. After reviewing process, we will make the pseudo stroke labels of all benchmarks available in the online repository.

\noindent
\textbf{CTW-1500}~\cite{liu2019curved} is a dataset for curve text detection which contains 1000 training and 500 testing images. The curved texts are labeled with 14 vertices at text-line level.

\noindent
\textbf{Total-Text}~\cite{ch2017total} consists of 1255 training and 300 testing images with a variety of text types including horizontal, multi-oriented and curved text instances. It is labeled with polygon and world-level annotations.

\noindent
\textbf{MSRA-TD500}~\cite{yao2012detecting} is a multi-language dataset including English and Chinese. There are 300 training images and 200 testing images and the text instances are annotated in the text-line level.

\noindent
\textbf{ICDAR 2015}~\cite{karatzas2015icdar} is another multi-oriented text detection dataset which includes 1000 training images and 500 testing images. The text areas are annotated as a quadrilateral.

\noindent
\textbf{ICDAR 2017 MLT}~\cite{nayef2017icdar2017} is a multi-oriented, multi-scripting and multi-lingual scene text dataset which consists of 7200 training images, 1800 validation images and 9000 testing images. The text areas are also annotated by four vertices of the quadrilateral.

\noindent
\textbf{ICDAR 2019 MLT}~\cite{nayef2019icdar2019} consists of 20,000 images containing text from 10 languages. The images are divided as follows: 50\% for training (a total of 10,000 images, 1,000 per language), and 50\% for testing. The text in the scene images of the dataset is annotated at word level, which is defined as a consecutive set of characters without spaces. Each text instance is labeled by a 4-corner bounding box associated with a script class as well as the corresponding Unicode transcription.

\begin{figure*}[h]
\includegraphics[width=.95\textwidth]{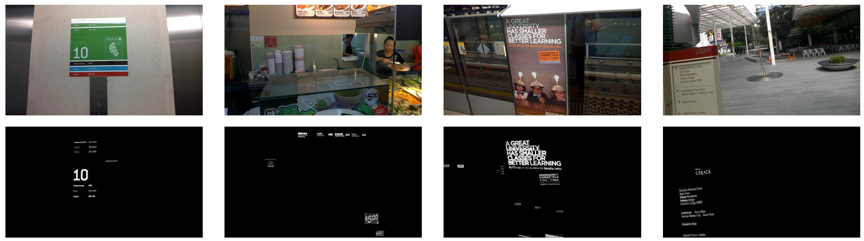}
\caption{Examples of benchmark datasets with extracted pseudo stroke labels.}
\label{fig:benchmark_example}
\end{figure*}

Fig.\ref{fig:benchmark_example} shows several examples sampled from benchmark datasets with their pseudo stroke labels. In the folder \textit{'Datasets/Stroke\_Labels\_Benchmark\_Datasets/ScreenShot'}, we show a partial screenshot of ICDAR 2015 and the corresponding stroke labels.

In addition, we introduce a dataset with stroke-level annotations, i.e., {\it SynthStroke}, to pre-train our StrokeNet before online detection. 
A link to download the entire dataset will be available in the online repository. After reviewing process, we will also make the code that generates the dataset available on the above website.
We expect the introduced stroke-level annotated dataset (which is very rare at present) can promote the research in text detection related areas. It has great potential to become general and popular dataset for pre-training in many fields such as text detection, image style and inpainting.

\subsection{SynthStroke}

{\it SynthStroke} is a \textbf{multi-orientated} and \textbf{multi-scale} text detection dataset, including a total of \textbf{800 thousand} synthetic images with the corresponding \textbf{stroke-level mask} and \textbf{text-level bounding box} annotations. 
The strokes of each synthetic image are labeled with the segmentation masks, while the text-level ground truth is annotated with word-level quadrangle.
To our best knowledge, it is the first synthetic dataset with stroke-level annotations in the field of text detection. 

\begin{figure*}[h]
\includegraphics[width=\textwidth]{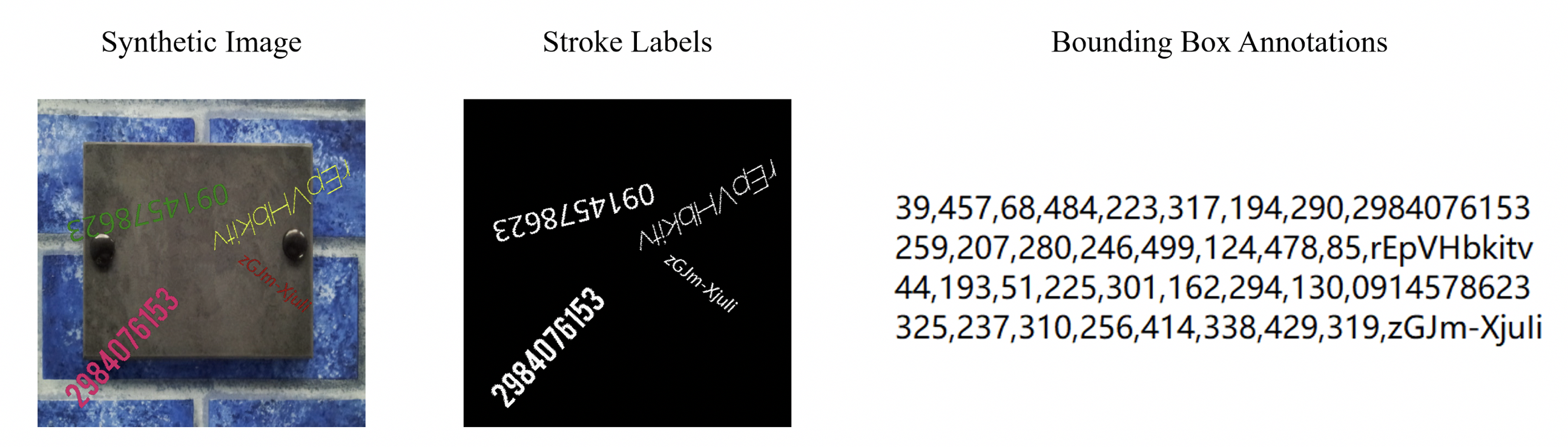}
\caption{An example of {\it SynthStroke} with annotations.}
\label{fig:synth_example}
\end{figure*}

\begin{figure*}
  \begin{center}
  \includegraphics[width=.85\textwidth]{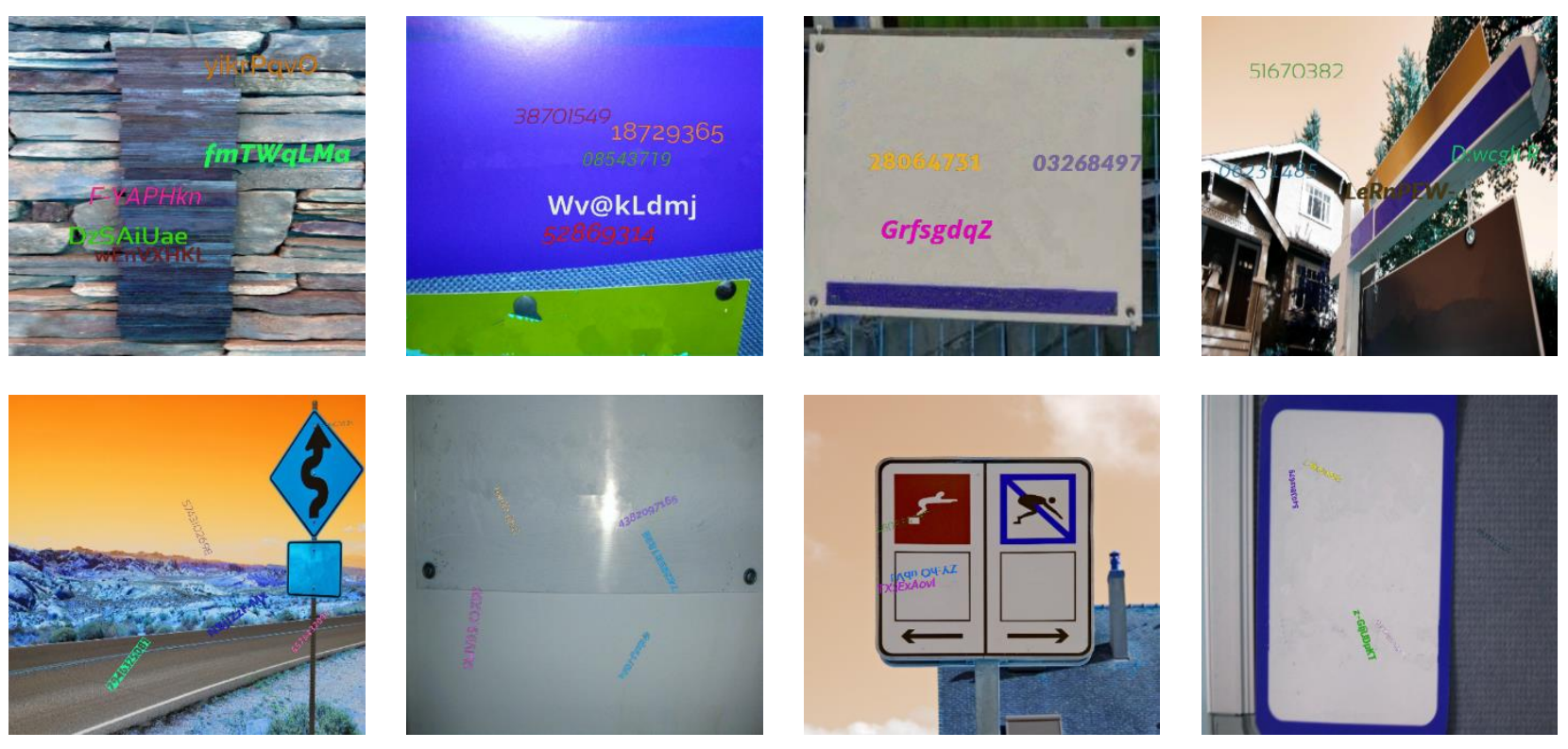}
  \caption{Synthetic examples of {\it SynthStroke}.}
  \label{fig:synth_examples}
  \end{center}
\end{figure*}

Fig.\ref{fig:synth_example} shows an example of our simulated dataset, while Fig.\ref{fig:synth_examples} visualizes several synthetic image examples. 
In the folder \textit{'Datasets/SynthStroke/ScreenShot'} of submitted supplementary materials, we show a partial screenshot of {\it SynthStroke} and the corresponding labels. 
Similarly, we provide a small number of samples of the dataset in the folder \textit{'Datasets/SynthStroke/Samples'}.

We synthesize the whole dataset based on the 8000 native images, which contain no texts and are collected from the open public repository \footnote{https://github.com/HCIILAB/Scene-Text-Removal}. 
It is necessary to note that another synthetic dataset, namely {\it SynthText}~\cite{gupta2016synthetic}, which is also synthesized from the above mentioned resource and commonly applied for the pre-training of many text detectors in previous research. Specifically, {\it SynthText} consists of approximately 800 thousand images with around 8 million synthetic word instances. Each text instance is annotated with its text-string, word-level and character-level bounding-boxes. A comprehensive comparison of {\it SynthStroke} and {\it SynthText} is summarized in Tabel \ref{tab-compar}. Compared to {\it SynthText}, our {\it SynthStroke} contains more diverse samples equipped with stroke labels, especially including some extremely small text instances which are helpful for training a more powerful text detector.
\begin{table*}[]
\begin{center}
\begin{tabular}{l|c|c}
\toprule
\textbf{Synthetic Dataset}   & \textbf{SynthStroke} & \textbf{SynthText} \\ \midrule
image quantity               & 800 thousand                & approximately 800 thousand               \\ \hline
text instance quantity       & around 8 million    & around 8 million   \\ \hline
text string                  & yes                  & yes                \\ \hline
word-level bounding box      & yes                  & yes                \\ \hline
character-level bounding box & no                   & yes                \\ \hline
stroke-level segmentation    & yes                  & no                 \\ \hline
font variation    & yes (100 types)                  & yes (less than 50 types)                 \\ \hline
orientation variation    & yes (from 0 to 360 degrees)                  & yes (from 0 to 180 degrees)                 \\ \hline
font size variation    & yes (from 5 to 80)                  & yes (from 20 to 50)                 \\ \hline
\end{tabular}
\end{center}
\caption{Comparison of our {\it SynthStroke} with {\it SynthText}.}
\label{tab-compar}
\end{table*}

\begin{figure*}
\begin{center}
\includegraphics[width=.75\textwidth]{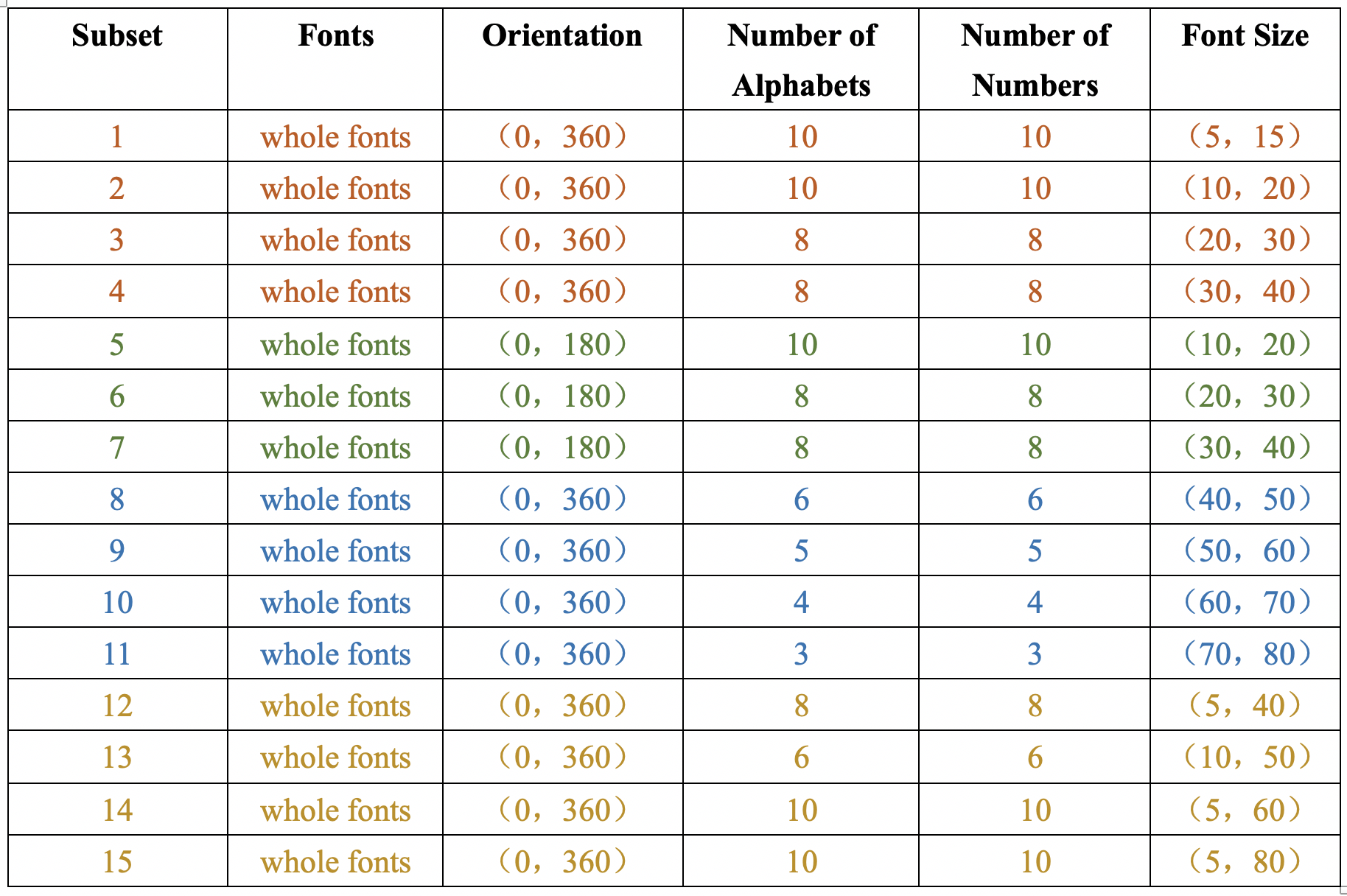}
\caption{Parameter configuration for dataset augmentations. For the subset one, it generates texts using the whole fonts which consst of 100 different types of fonts. Besides, the rotation angle of the text instance is randomly selected between 0 and 360 degrees, while the maximum length of each text instance is 20, containing 10 alphabets and 10 numbers. Moreover, the range of font size is set from 5 to 15.}
\label{fig:configs}
\end{center}
\end{figure*}

Since {\it SynthText} has maintained a relatively conservative variation of text attributes such as font and rotation angle during the synthesis process, which limits the diversity of text forms. In contrast, we perform augmentations of dataset to ensure the diversity of the whole {\it SynthStroke}. Concretely, we apply different configuration of parameters through the image synthesis, including font, font size, rotation angle, the number of alphabets and numbers, to control the generation of training samples. In the meantime, we divide the whole dataset into 15 subsets to meet the needs of training data in different scenarios. The detailed parameter configurations of subsets are highlighted in Fig.\ref{fig:configs}.

There are 100 different types of fonts, and the whole fonts set includes
\begin{scriptsize} 
\textit{'AllerDisplay.ttf', 'Aller\_Bd.ttf', 'Aller\_BdIt.ttf', 'Aller\_It.ttf', 'Aller\_Lt.ttf', 'Aller\_LtIt.ttf', 'Aller\_Rg.ttf', 'Amatic-Bold.ttf', 'AmaticSC-Regular.ttf', 'BEBAS\_\_\_.ttf', 'Capture\_it.ttf', 'Capture\_it\_2.ttf', 'CaviarDreams.ttf', 'CaviarDreams\_BoldItalic.ttf', 'CaviarDreams\_Italic.ttf', 'Caviar\_Dreams\_Bold.ttf', 'DroidSans-Bold.ttf', 'DroidSans.ttf', 'FFF\_Tusj.ttf', 'Lato-Black.ttf', 'Lato-BlackItalic.ttf', 'Lato-Bold.ttf', 'Lato-BoldItalic.ttf', 'Lato-Hairline.ttf', 'Lato-HairlineItalic.ttf', 'Lato-Heavy.ttf', 'Lato-HeavyItalic.ttf', 'Lato-Italic.ttf', 'Lato-Light.ttf', 'Lato-LightItalic.ttf', 'Lato-Medium.ttf', 'Lato-MediumItalic.ttf', 'Lato-Regular.ttf', 'Lato-Semibold.ttf', 'Lato-SemiboldItalic.ttf', 'Lato-Thin.ttf', 'Lato-ThinItalic.ttf', 'OpenSans-Bold.ttf', 'OpenSans-BoldItalic.ttf', 'OpenSans-ExtraBold.ttf', 'OpenSans-ExtraBoldItalic.ttf', 'OpenSans-Italic.ttf', 'OpenSans-Light.ttf', 'OpenSans-LightItalic.ttf', 'OpenSans-Regular.ttf', 'OpenSans-Semibold.ttf', 'OpenSans-SemiboldItalic.ttf', 'Pacifico.ttf', 'Raleway-Black.ttf', 'Raleway-BlackItalic.ttf', 'Raleway-Bold.ttf', 'Raleway-BoldItalic.ttf', 'Raleway-ExtraBold.ttf', 'Raleway-ExtraBoldItalic.ttf', 'Raleway-ExtraLight.ttf', 'Raleway-ExtraLightItalic.ttf', 'Raleway-Italic.ttf', 'Raleway-Light.ttf', 'Raleway-LightItalic.ttf', 'Raleway-Medium.ttf', 'Raleway-MediumItalic.ttf', 'Raleway-Regular.ttf', 'Raleway-SemiBold.ttf', 'Raleway-SemiBoldItalic.ttf', 'Raleway-Thin.ttf', 'Raleway-ThinItalic.ttf', 'Roboto-Black.ttf', 'Roboto-BlackItalic.ttf', 'Roboto-Bold.ttf', 'Roboto-BoldItalic.ttf', 'Roboto-Italic.ttf', 'Roboto-Light.ttf', 'Roboto-LightItalic.ttf', 'Roboto-Medium.ttf', 'Roboto-MediumItalic.ttf', 'Roboto-Regular.ttf', 'Roboto-Thin.ttf', 'Roboto-ThinItalic.ttf', 'RobotoCondensed-Bold.ttf', 'RobotoCondensed-BoldItalic.ttf', 'RobotoCondensed-Italic.ttf', 'RobotoCondensed-Light.ttf', 'RobotoCondensed-LightItalic.ttf', 'RobotoCondensed-Regular.ttf', 'Sansation-Bold.ttf', 'Sansation-BoldItalic.ttf', 'Sansation-Italic.ttf', 'Sansation-Light.ttf', 'Sansation-LightItalic.ttf', 'Sansation-Regular.ttf', 'SEASRN\_\_.ttf', 'Walkway\_Black.ttf', 'Walkway\_Bold.ttf', 'Walkway\_Oblique.ttf', 'Walkway\_Oblique\_Black.ttf', 'Walkway\_Oblique\_Bold.ttf', 'Walkway\_Oblique\_SemiBold.ttf', 'Walkway\_Oblique\_UltraBold.ttf', 'Walkway\_SemiBold.ttf', 'Walkway\_UltraBold.ttf'}.
\end{scriptsize} 

We introduce {\it SynthStroke} to pre-train the whole pipeline of our StrokeNet, which is helpful for subsequent performing fine-tuning and evaluation on real scene datasets.

\section{Code and Implementations}

In this section, we provide a brief introduction to the implementation details of the experiments on benchmark datasets. Experiments were conducted on NVIDIA GeForce GTX 1080Ti GPU and implemented in Pytorch. The backbone was pre-trained on ImageNet. 
We adopted the data augmentation strategy that input images were resized to 640*640, and each image was randomly flipped with a probability of 0.5. The training batch size was set to 16, with 4 images per GPU. We adopted Adam \cite{kingma2014adam} as the optimizer with the momentum 0.9 and the learning rate 10-4 for pre-training, and used SGD as the optimizer with the learning rate 0.03 and decay rate 0.5 per 100 epochs for fine-tuning. The Pytorch implementation will be available.
We will only highlight the important tricks that can help improve the model performance below.

\subsection{Close and arbitrary-shaped text detection}

To evaluate the performance of our StrokeNet for detecting close or arbitrary-shaped text instances, we compare the proposed model with several state-of-the-art methods on two curved benchmarks, named Total-Text and CTW-1500. Since it is challenging to directly train the model on these datasets because their annotations are obtained by complicating text area cropping for splitting character boxes during weakly-supervised learning, ICDAR 2015 dataset was used to pre-train our model and fine-tuning was then conducted on above two datasets, separately. The longer sides of the images within Total-Text and CTW-1500 were resized to 1280 and 1024, respectively. The quantitative results are listed in Table 1, while the visualization of curved text detection results are shown in Figure 6 of the submitted manuscript.

\subsection{Small and low-resolution text detection}

We evaluate our method on ICDAR 2015 to validate its ability for detecting multi-oriented texts. This dataset also contains a lot of small and low-resolution text instances. Similar to previous methods, the model was evaluated with the original image size of 720*1280. The quantitative results are listed in Table 1, while the visualization of multi-oriented text detection results are shown in Figure 6 of the submitted manuscript.

\subsection{Multi-language text detection}

To test the robustness of StrokeNet to multiple languages with long texts, we evaluate our method on MSRA-TD500, ICDAR 2017 MLT and ICDAR 2019 MLT benchmarks. To ensure fair comparisons, we resized the short edge of test images to 512 if it is less than 512, and kept the long edge is not larger than 2048. The quantitative results are listed in Table 2, while the qualitative results are shown in Figure 6 of the submitted manuscript.

\subsection{Loss Functions}
\begin{align}
\begin{array}{l}
{\begin{array}{*{20}{l}}{L\mathop{{}}\nolimits_{{sin}}=\frac{{1}}{{ \left| N \right| }}{\mathop{ \sum }\limits_{{i \in N}}{smooth\mathop{{}}\nolimits_{{L\mathop{{}}\nolimits_{{1}}}} \left( sin \theta \mathop{{}}\nolimits_{{i}}-sin\mathop{{ \theta }}\limits^{ \sim }\mathop{{}}\nolimits_{{i}} \right) }}} , \nonumber \\{L\mathop{{}}\nolimits_{{cos}}=\frac{{1}}{{ \left| N \right| }}{\mathop{ \sum }\limits_{{i \in N}}{smooth\mathop{{}}\nolimits_{{L\mathop{{}}\nolimits_{{1}}}} \left( cos \theta \mathop{{}}\nolimits_{{i}}-cos\mathop{ \theta }\limits^{ \sim }\mathop{{}}\nolimits_{{i}} \right) }}} , \nonumber \end{array}}
\end{array}
\end{align}
where ${N}$ represents pixels set and ${ \left| N \right| }$ means the number of pixels in TCA. 
${sin \theta }$ and ${cos \theta }$ are the predicted angle values, while ${sin\mathop{{ \theta }}\limits^{ \sim }}$ and ${cos\mathop{{ \theta }}\limits^{ \sim }}$ are the corresponding ground-truth labels. 
Note that $smooth_{L_1}$ loss~\cite{ren2016faster} is a modified $L_1$ loss with smooth gradient near zero. 

\begin{align}
L_h = \frac{1}{2\left|N\right|}\sum_{i\in N}\sum_{k=1}^2 \left(\log(h+1) * smooth_{L_1}\left(\frac{h_{ki}}{\tilde{h}_ki} - 1\right)\right), \nonumber
\end{align}
where the weight ${\log\text{(}h+1\text{)}}$ is introduced to promote the regression for text instance with large scale height ${h}$.

\begin{align}
L_{MSE} = \frac{1}{|N|} \sum_{i \in N} \left(\alpha_i^p - \alpha_i^g \right)^2, \nonumber
\end{align}
where ${\mathop{{ \alpha }}\nolimits_{{i}}^{{p}}}$ and ${\mathop{{ \alpha }}\nolimits_{{i}}^{{g}}}$ are the predicted and ground truth stroke values at pixel ${i}$ respectively. 

\begin{align}
L_{SSIM} = 1 - \frac{(2\mu_p\mu_g + c_1)(2\sigma_{pg} + c_2)}{(\mu_2^p + \mu_2^g + c_1)(\sigma_2^p + \sigma_2^g + c_2)}, \nonumber
\end{align}
where ${\mathop{{ \mu }}\nolimits_{{p}}}$, ${\mathop{{ \mu }}\nolimits_{{g}}}$ and ${\mathop{{ \sigma }}\nolimits_{{p}}}$, ${\mathop{{ \sigma }}\nolimits_{{g}}}$ are the mean and standard deviations of ${\mathop{{ \alpha }}\nolimits_{{i}}^{{p}}}$ and ${\mathop{{ \alpha }}\nolimits_{{i}}^{{g}}}$.
\begin{figure*}
\includegraphics[width=\textwidth]
{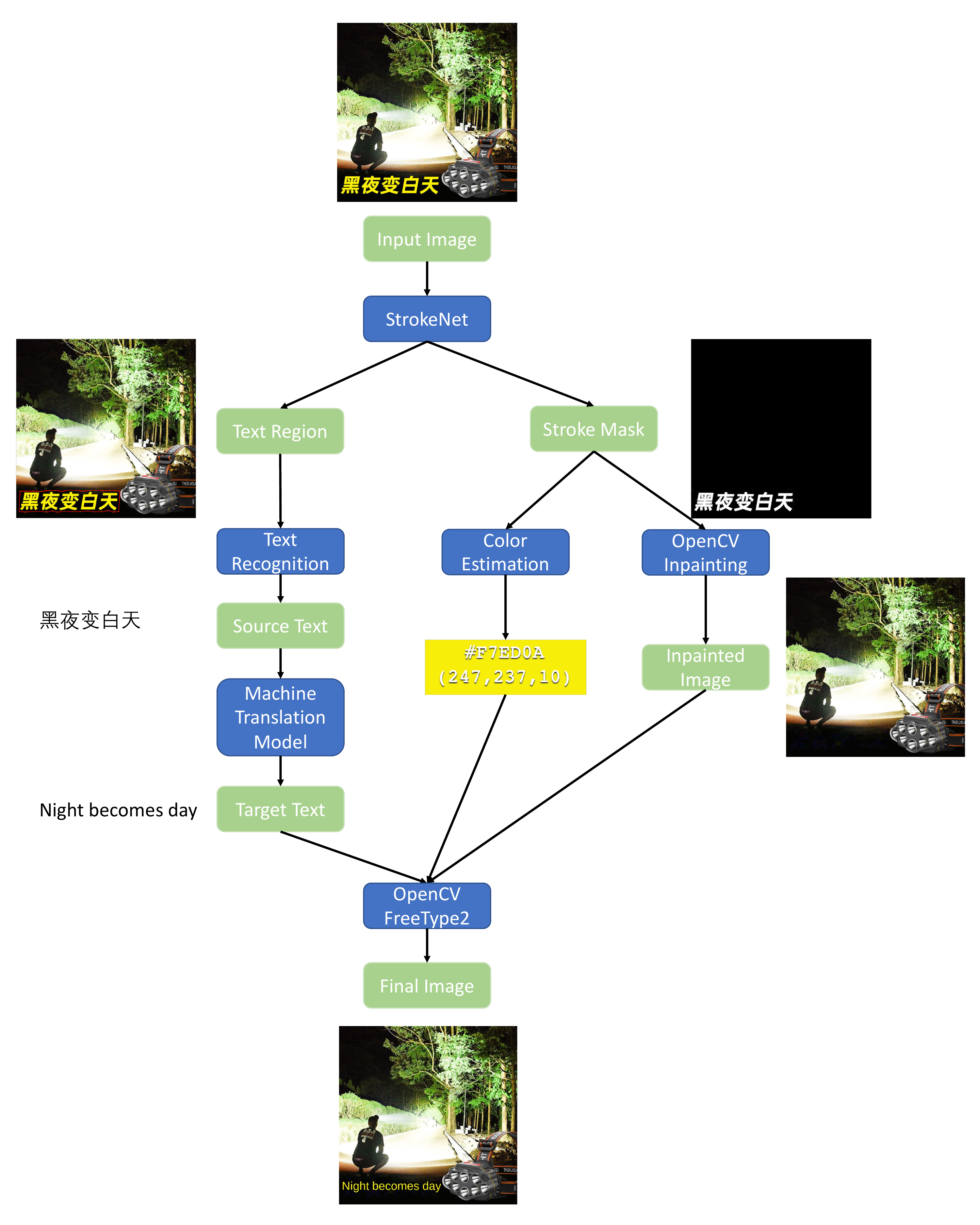}
\caption{The pipeline of OCT translation with StrokeNet. An example of Chinese to English OCR translation is illustrated.}
\label{fig:ocr_trans}
\end{figure*}

\begin{figure*}
\centering

        \includegraphics[width=0.4\textwidth]{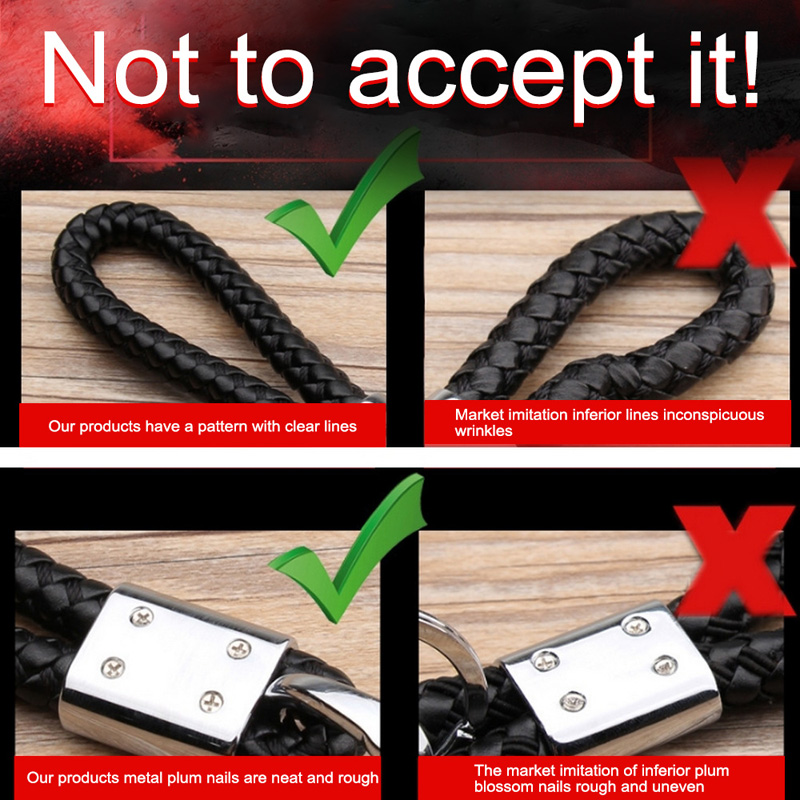}
        \includegraphics[width=0.4\textwidth]{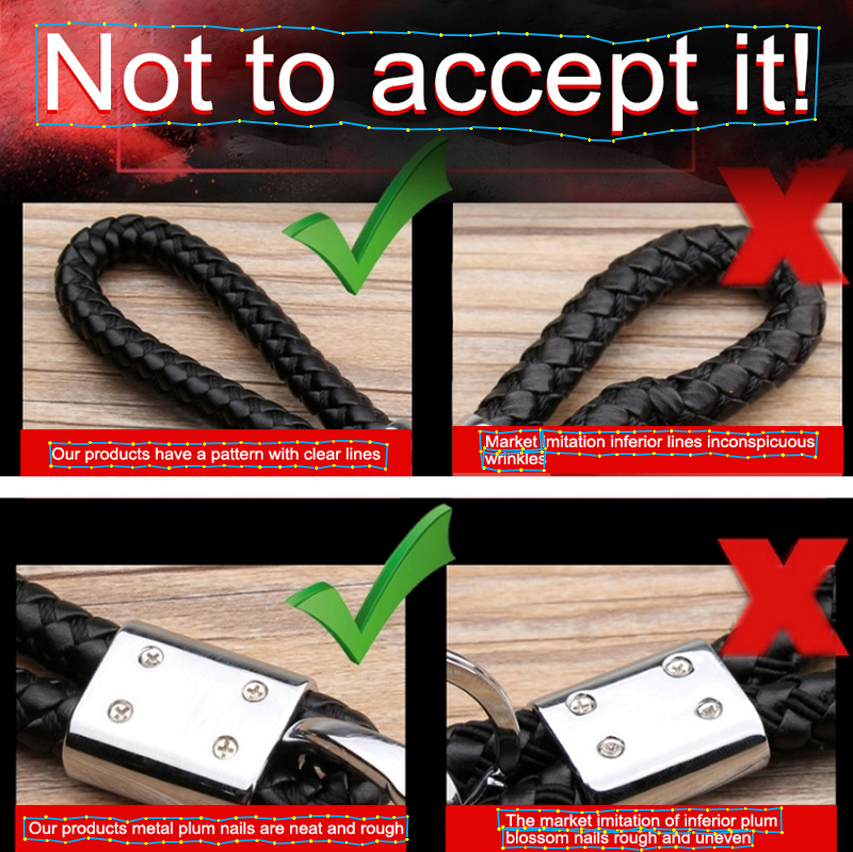}
        \includegraphics[width=0.4\textwidth]{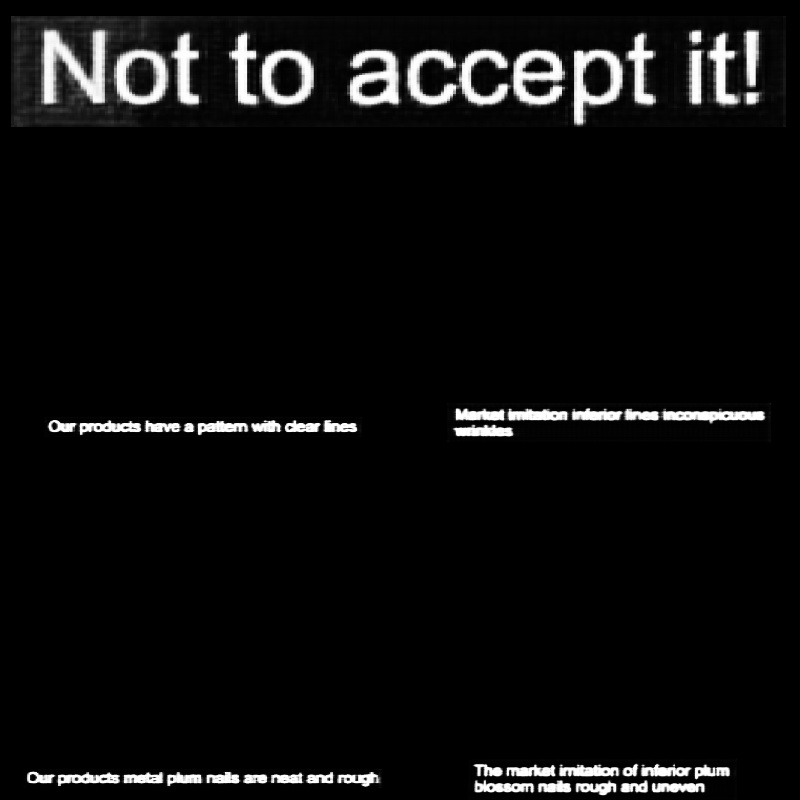}
        \includegraphics[width=0.4\textwidth]{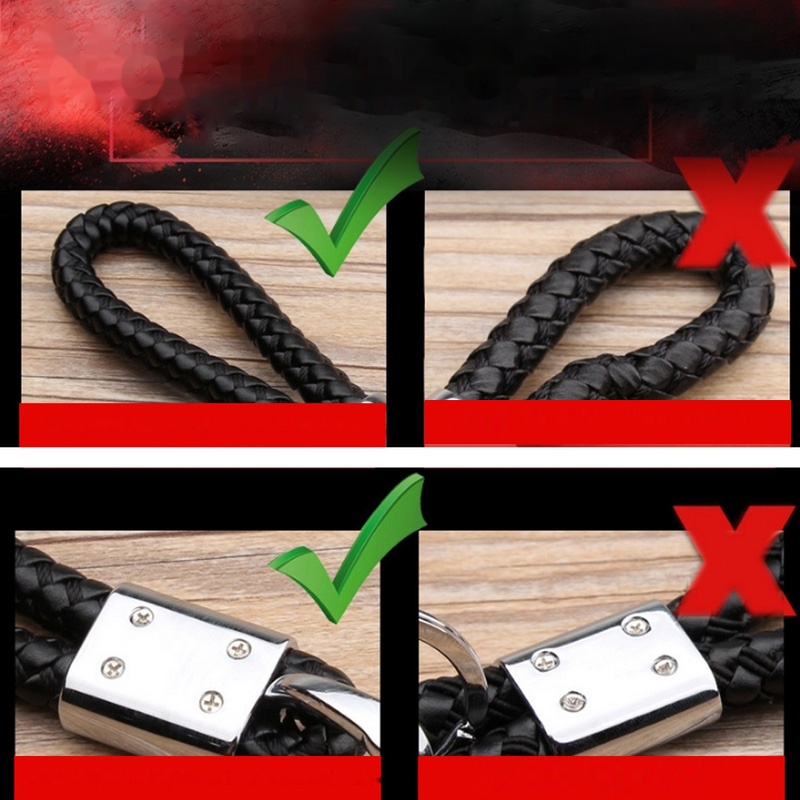}
        \includegraphics[width=0.4\textwidth]{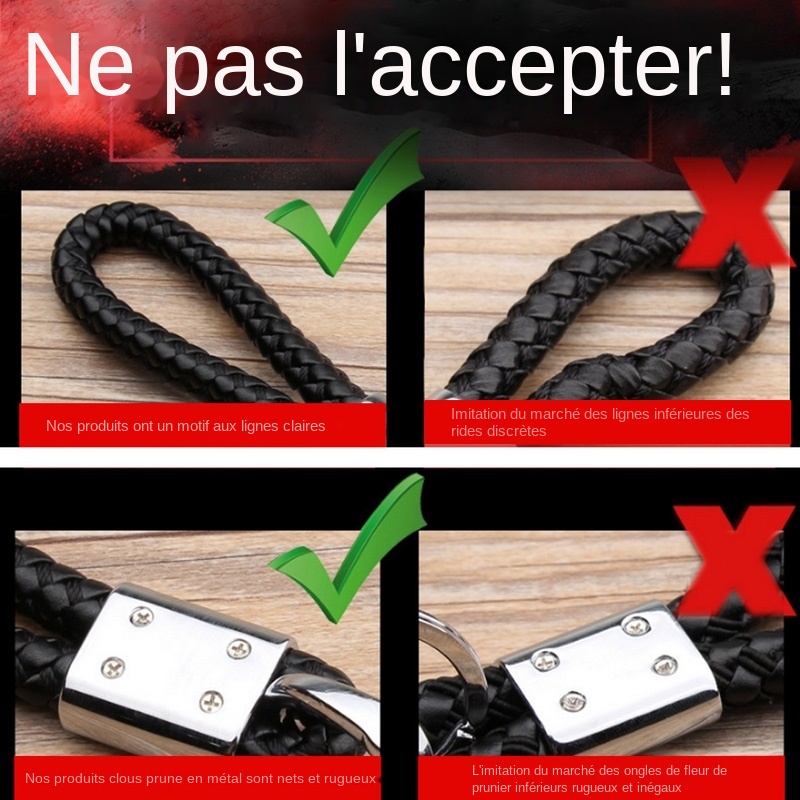}
\caption{An example of OCR translation from English to French. From top-to-bottom and left-to-right, the figures represent \textbf{Input image}, \textbf{Text detection result}, \textbf{Predicted stroke-level mask}, \textbf{Inpainted image} and \textbf{Translated image}.}
\label{fig:ocr_trans_example}
\end{figure*}
\subsection{Graph Generation}
A isomorphic stroke graph and a heterogeneous text graph are built separately for hierarchical relations reasoning and linkages prediction by extracting candidate bounding boxes from both levels.
Specifically, we treat each stroke-level proposal as a node and select a limited number of its neighbors to construct multiple isomorphic graphs. For a center node, we adopt its neighbors within ${2}$ hops to generate local structure. In our setting, ${1}$-hop of a center node contains ${8}$ nearest neighbors, while its ${2}$-hop includes ${4}$ nearest neighbors. The limited number of neighbors is useful for effectively relational reasoning, while high-order neighbors providing auxiliary information of the local structure. After getting the center location of each box, we employ KNN operation~\cite{zhang2020deep} to build isomorphic stroke graphs based on the Euclidean distance. After that, we apply similar technique to text-level proposals but connect each text node with specified number of stroke nodes if the area of the former contains the center of the latter, to further build a few heterogeneous text graphs which are composed of nodes at both levels.

\section{Application: OCR Translation}

In this section, we describe the application of OCR translation as the downstream task of our StrokeNet model. 
Fig.\ref{fig:ocr_trans} shows the overall pipeline of the proposed application. 
Fig.\ref{fig:ocr_trans_example} shows the detailed example in the submitted manuscript.

The StrokeNet is first called to output the stroke- and text-level detection results. 
They will be separately fed into a text recognition model and an inpainting module. 
The text recognition model is our in-house toolkit, which is developed for internal usage. 
As the image inpainting method, we simply use the built-in OpenCV inpaint algorithm\footnote{https://docs.opencv.org/master/df/d3d/tutorial\_py\_inpainting.html}. 
Note that before applying the inpainting operation, we apply a dilation operation with the built-in OpenCV function to enlarge the stroke masked areas. 
In addition, with the stroke mask, we can readily estimate the text color by averaging the pixel values in masking region. 
This strategy can in practice improve the visual effect of inpainting. 
The machine translation model is run by calling Google translate API \footnote{https://translate.google.com/}, including a language identification API. 

\end{document}